\pdfoutput=1

\documentclass[11pt]{article}

\usepackage[]{acl}

\definecolor{blue1}{HTML}{d1eeea}
\definecolor{blue2}{HTML}{a8dbd9}
\definecolor{blue3}{HTML}{85c4c9}
\definecolor{blue4}{HTML}{68abb8}
\definecolor{blue5}{HTML}{4f90a6}
\definecolor{blue6}{HTML}{3b738f}
\definecolor{blue7}{HTML}{2a5674}

\newcommand{\gradecolor}[1]{%
  \ifnum#1<30 \cellcolor{blue1}\color{black}%
  \else \ifnum#1<40 \cellcolor{blue2}\color{black}%
  \else \ifnum#1<50 \cellcolor{blue3}\color{black}%
  \else \ifnum#1<58 \cellcolor{blue4}\color{black}%
  \else \ifnum#1<66 \cellcolor{blue5}\color{white}%
  \else \ifnum#1<78 \cellcolor{blue6}\color{white}%
  \else \cellcolor{blue7}\color{white}%
  \fi\fi\fi\fi\fi\fi%
}

\usepackage{times}
\usepackage{latexsym}
\usepackage{subcaption}
\usepackage{lipsum}  %

\usepackage[T1]{fontenc}

\usepackage[utf8]{inputenc}

\usepackage{microtype}

\usepackage{inconsolata}

\usepackage{graphicx}

\usepackage{times}
\usepackage{latexsym}
\usepackage{lipsum}     %
\usepackage{ntheorem}   %
\usepackage{mdframed}   %
\usepackage{csquotes}
\theoremstyle{break}
\theoremheaderfont{\bfseries}
\newmdtheoremenv[%
linecolor=gray,leftmargin=60,%
rightmargin=40,
backgroundcolor=gray!40,%
innertopmargin=0pt,%
ntheorem]{myprop}{Proposition}[section]

\definecolor{myblue}{HTML}{349BFB} %
\definecolor{mygreen}{HTML}{35978f} %

\usepackage{pdfrender}

\usepackage[T1]{fontenc}

\usepackage{times}
\usepackage{latexsym}
\usepackage{amsmath,graphicx}
\usepackage{bm}
\usepackage{multirow}
\usepackage{url}
\usepackage{rotating}
\usepackage{pdflscape}
\usepackage{comment}
\usepackage{amssymb}
\usepackage{stfloats}
\usepackage{tabu}
\usepackage{bbm}
\usepackage{array}
\usepackage{hyperref}
\usepackage{footnote}
\usepackage{booktabs}
\usepackage[flushleft]{threeparttable}
\makesavenoteenv{tabular}
\makesavenoteenv{table}
\usepackage{tablefootnote}
\usepackage{enumitem}
\usepackage{amssymb}%
\usepackage{pifont}%
\usepackage{makecell}
\usepackage{threeparttable}
\usepackage{soul}
\usepackage[normalem]{ulem}
\usepackage{tabu}

\definecolor{colorAdd}{HTML}{FFC2BA}

\makeatletter
\newcommand{\hlll}[1]{%
  \bgroup
  \def\tempcolor{#1}%
  \UL@protected\def\sout{\bgroup \ULdepth =-.8ex \ULset}%
  \markoverwith{\textcolor[HTML]{\tempcolor}{\rule[-.5ex]{.1pt}{2.5ex}}}%
  \ULon}
\makeatother

\makeatletter
\newcommand\hll{%
  \bgroup
  \UL@protected\def\sout{\bgroup \ULdepth =-.8ex \ULset}%
  \markoverwith{\textcolor{colorAdd}{\rule[-.5ex]{.1pt}{2.5ex}}}%
  \ULon}
\makeatother

\usepackage{pgf} %
\usepackage{colortbl}

\definecolor{LightColor}{HTML}{CBEBE7}
\definecolor{DarkColor}{HTML}{264B68}

\usepackage[T1]{fontenc}
\usepackage{adjustbox}

\usepackage[utf8]{inputenc}

\usepackage{microtype}

\newcolumntype{L}[1]{>{\raggedright\arraybackslash}p{#1}}
 
\newcolumntype{C}[1]{>{\centering\arraybackslash}p{#1}}
 
\newcolumntype{R}[1]{>{\raggedleft\arraybackslash}p{#1}}

\usepackage{inconsolata}

\usepackage[colorinlistoftodos]{todonotes}

\newcommand{\medcsi}{\textsc{MedReadMe}}

\title{\medcsi: A Systematic Study for Fine-grained\\Sentence Readability in Medical Domain}

\author{Chao Jiang\\
  College of Computing\\Georgia Institute of Technology \\
  chaojiang@gatech.edu \And
  Wei Xu\\
  College of Computing\\Georgia Institute of Technology \\
  wei.xu@cc.gatech.edu
}

\begin{document}
\maketitle

\begin{abstract}
  Medical texts are notoriously challenging to read. Properly measuring their readability is the first step towards making them more accessible. In this paper, we present a systematic study on fine-grained readability measurements in the medical domain at both sentence-level and span-level. We introduce a new dataset \medcsi, which consists of manually annotated readability ratings and fine-grained complex span annotation for 4,520 sentences, featuring two novel ``Google-Easy'' and ``Google-Hard'' categories. It supports our quantitative analysis, which covers 650 linguistic features and automatic complex word and jargon identification. Enabled by our high-quality annotation, we benchmark and improve several state-of-the-art sentence-level readability metrics for the medical domain specifically, which include unsupervised, supervised, and prompting-based methods using recently developed large language models (LLMs). Informed by our fine-grained complex span annotation, we find that adding a single feature, capturing the number of jargon spans, into existing readability formulas can significantly improve their correlation with human judgments. The data is available at \url{tinyurl.com/medreadme-repo}.

\end{abstract}

\section{Introduction}

\begin{quote}
  \textit{If you can’t measure it, you can’t improve it.}
  
    \vspace{-8pt}
    
  \hfill -- Peter Drucker
\end{quote}

\vspace{-4pt}

\begin{figure}[t!]
    \centering
    \includegraphics[width=\linewidth]{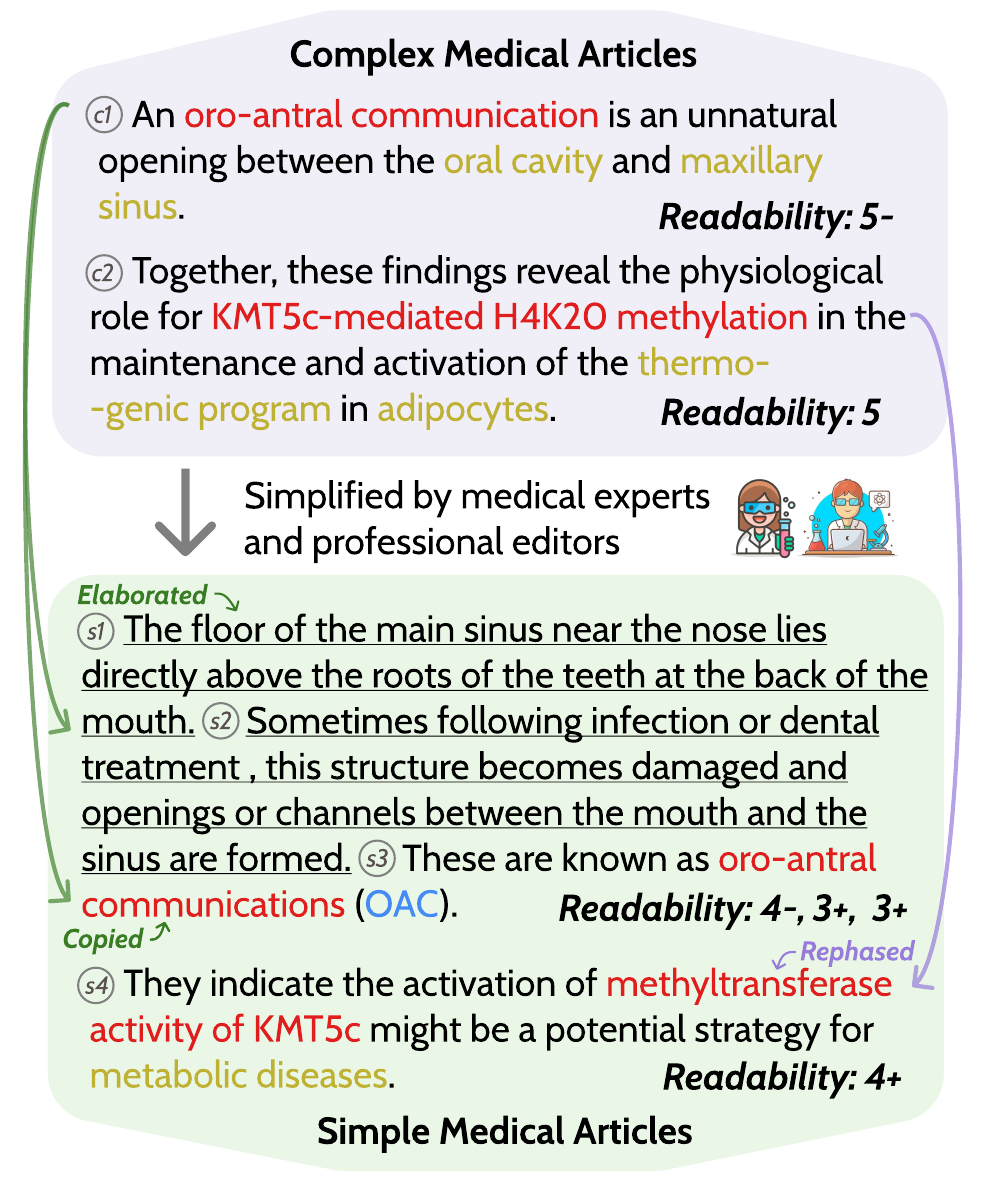}
    \vspace{-20pt}
    \caption{An illustration of our dataset, with sentence readability ratings and fine-grained complex span annotation on 4,520 sentences, including \textcolor[HTML]{E11E1E}{``Google-Hard''} and \textcolor[HTML]{BFB133}{``Google-Easy''}, \textcolor[HTML]{448BF5}{abbreviations}, and general complex terms, etc. We also analyze how medical jargon are being handled during simplification. e.g., a Google-Hard \textcolor[HTML]{E11E1E}{``oro-antral communication''} is copied and elaborated. Some jargon are ignored for clarity.}
    \vspace{-20pt}
    \label{figure:page1}
\end{figure}

\noindent  Timely disseminating reliable medical knowledge to those in need is crucial for public health management \cite{august2023paper}. Trustworthy platforms like Merck Manuals and Wikipedia contain extensive medical information, while research papers introduce the latest findings, including emerging medical conditions and treatments \cite{joseph-etal-2023-multilingual}. However, comprehending these resources can be very challenging due to their technical nature and the extensive use of specialized terminology \cite{zeng2005text}.
As the first step to making them more accessible, properly measuring the readability of medical texts is crucial \cite{rooney2021readability,echuri2022readability}. However, a high-quality multi-source dataset for reliably evaluating and improving sentence readability metrics for medical domain is lacking.

To address this gap in research, we present a systematic study for medical text readability in this paper, which includes a manually annotated readability dataset (\S \ref{section:dataset-construction}), a data-driven analysis to answer \textit{``why medical sentences are so hard''}, covering 650 linguistic features and additional medical jargon features (\S \ref{sec:key-findings}), a comprehensive benchmark of state-of-the-art readability metrics (\S \ref{section:evaluating-existing-readability}), a simple yet effective method to improve LM-based readability metrics by training on our dataset (\S \ref{section:add-jar-method}), and an automatic model that can identify complex words and jargon with fine-grained categories (\S \ref{sec:cwi_experiment}). 

\begin{figure*}[pht!]
    \centering
    \vspace{-21pt}
    \includegraphics[width=\linewidth]{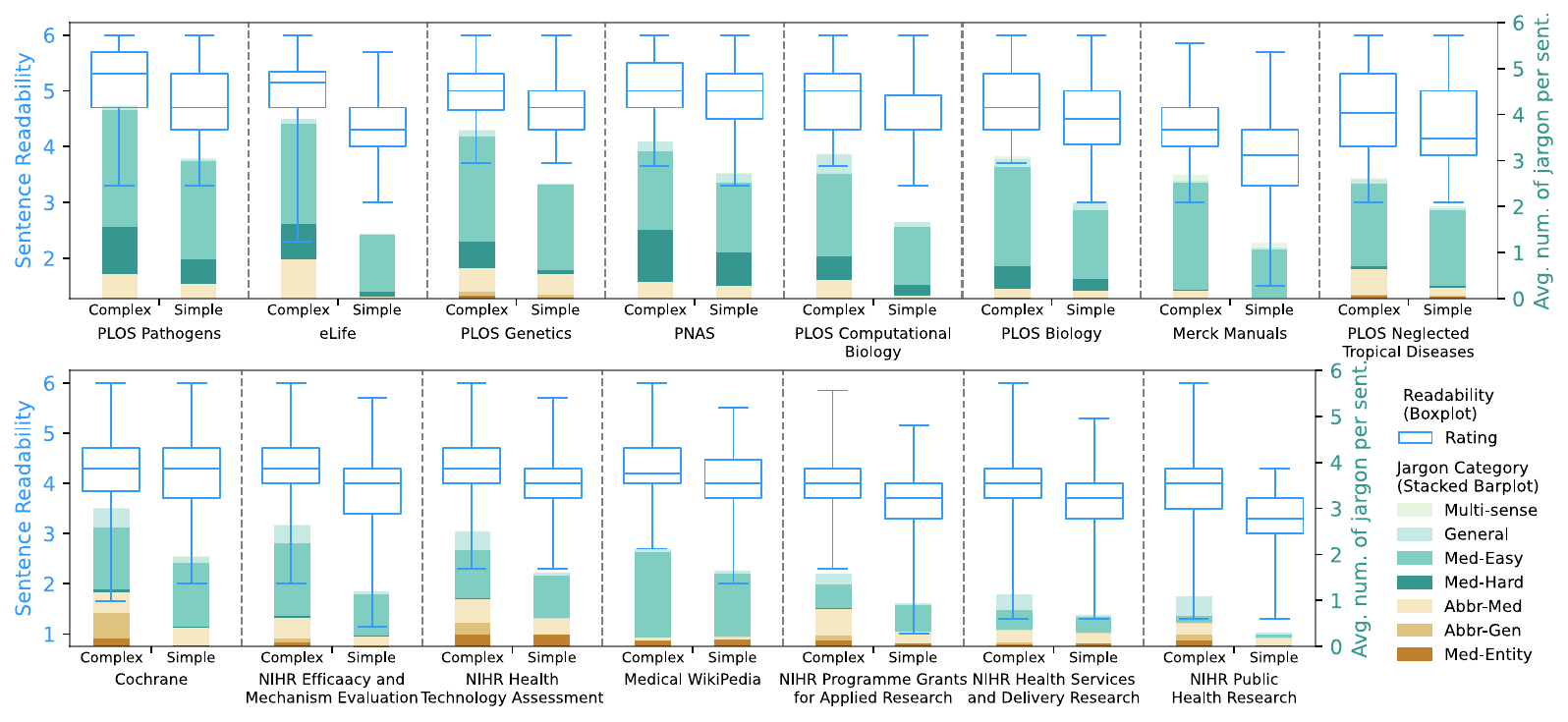}
    \vspace{-19pt}
    \caption{The distribution of sentence readability \textcolor{myblue}{(boxplot on the left y-axis)} and the average number of jargon spans per category \textcolor{mygreen}{(stacked barplot on the right y-axis)} in each sentence across both ``complex'' and ``simplied'' versions for 15 commonly used resources for medical text simplification. \textcolor{myblue}{Sentences with higher readability scores require a higher level of education to comprehend.} The readability of sentences in different resources varies greatly.}
    \label{fig:readability-jargon-distribution}
    \vspace{-17pt}
\end{figure*}

Our \medcsi~ dataset consists of 4,520 sentences with both sentence-level readability ratings and fine-grained complex span-level annotations (Figure \ref{figure:page1}). It covers complex-simple parallel article pairs from 15 diverse data resources that range from encyclopedias to plain-language summaries to biomedical research publications (Figure \ref{fig:readability-jargon-distribution}).   The readability ratings are annotated using a rank-and-rate interface \cite{maddela-etal-2023-lens} based on the CEFR scale \cite{arase-etal-2022-cefr}, which is shown to be more reliable than other methods \cite{naous2023towards}. We also ask lay annotators to highlight any words/phrases that they find hard to understand and categorize the reason using a 7-class taxonomy.  Considering that ``the majority of people seek health information online began at a search engine'',\footnote{\url{https://tinyurl.com/seek-health-info-online}} we introduce two categories of ``Google-Easy'' and ``Google-Hard'' to reflect whether jargon is understandable after a quick Google search, providing a fresh perspective beyond binary or 5-point Likert scales.

Our new dataset addresses three limitations in prior work: (1) Existing work with sentence-level ratings mainly covers data from general domains, such as Wikipedia \cite{de-clercq-hoste-2016-mixed}, news \cite{vstajner2017automatic, brunato-etal-2018-sentence}, and textbooks for ESL learners \cite{arase-etal-2022-cefr}, which are very different from specialized fields, such as medicine \cite{choi2007multidisciplinarity}. (2) Prior work separates the research on sentence readability and complex jargon terms, hence missing the possible correlations between them \cite{kwon-etal-2022-medjex, naous2023towards}. (3) Previous research on sentence readability uses document-level ratings as an approximation, which is shown to be inaccurate \cite{arase-etal-2022-cefr, cripwell2023simplicity}.

Our analysis reveals that compared to various linguistic features, complex spans, especially medical jargon from certain domains, more significantly elevate the difficulty of sentences (\S \ref{sec:why-medical-sentences-are-hard}). We also scrutinize the quality of  15 widely used medical text simplification resources (\S \ref{section:variance-of-readability}), and find that there are non-negligible variances in readability among them, as shown by the differences in the height of the box plots in Figure \ref{fig:readability-jargon-distribution}. While evaluating various sentence readability metrics, we find that unsupervised methods based on lexical features perform poorly in the medical domain. Prompting large language models such as GPT-4 \cite{achiam2023gpt} with 5-shot achieves strong performance, yet is outperformed by fine-tuned models in a much smaller size. 
Inspired by our analysis, we add a single feature that captures the  ``number of jargon'' in a sentence into existing readability formulas, and find it can significantly improve their performance and also make them more stable.

\section{Constructing \textsc{\medcsi} Corpus}
\label{section:dataset-construction}

This section presents the detailed procedure for constructing the Medical Readability Measurement (\textsc{\medcsi}) corpus, which consists of  4,520 sentences in 180 complex-simple article pairs randomly sampled from 15 data sources (\S \ref{section:data-collection}).

\renewcommand{\arraystretch}{1.1}
\begin{table*}[t!]
    \centering
    \vspace{-10pt}
    \resizebox{\linewidth}{!}{
\begin{tabular}{@{\hspace{0.02cm}}L{3.0cm}@{\hspace{0.2cm}}L{7.6cm}@{\hspace{0.2cm}}l@{\hspace{0.05cm}}c@{\hspace{0.05cm}}r@{\hspace{0.02cm}}}
\toprule
    \textbf{Category}     & \textbf{Definition}  & \textbf{Example}  & \textbf{Tok. Len.} & \textbf{\%} \\ \midrule
       \multicolumn{3}{@{\hspace{0.02cm}}l}{\textbf{Medical Jargon} } & 2.2$\pm$1.5 & 68.6\% \\ \midrule
        \quad Google-Easy  & \makecell[l]{Medical terms that can be easily understood\\after a quick search.} & \makecell[l]{\hll{Schistosoma mansoni} is a parasitic infection\\common in the tropics and sub-tropics.} & 2.0$\pm$1.2 & 56.9\% \\  \midrule
        \quad Google-Hard & \makecell[l]{Medical terms that require extensive research\\before a layperson can possibly understand them.} & \makecell[l]{\dots retains limited DNA-processing activity,\\albeit via a \hll{distributive binding mechanism}.} & 3.2$\pm$2.5 & 7.5\% \\  \midrule
        \quad Name Entity & \makecell[l]{Brand or organization name, excluding general\\medical terms such as drugs and equipments.} & \makecell[l]{While vaccination with \hll{BioNTech} and \hll{Moderna}\\mostly causes only mild and typical \dots} & 2.7$\pm$2.2 & 4.1\% \\ \midrule
        \textbf{General Complex} & \makecell[l]{Terms that are outside the vocabulary of 10-12th\\graders and not specific to the medical domain.} & \makecell[l]{Treatments used to \hll{ameliorate} symptoms and\\reduce morbidity include opiates, sedatives \dots} & 1.9$\pm$1.2 & 10.2\% \\  
        \midrule
        \textbf{Multi-sense} & \makecell[l]{Spans that have different meanings in the\\medical context compared to their general use.} & \makecell[l]{\dots in structural and/or functional aspects of the\\interaction with the insect \hll{vector}.} & 1.0$\pm$0.1 & 0.5\% \\
        \midrule
        \multicolumn{3}{@{\hspace{0.02cm}}l}{\textbf{Abbreviation} } & 1.1$\pm$0.4 & 20.8\% \\ \midrule
        \quad Medical Domain & \makecell[l]{Abbreviations that have a specific meaning in\\the medical domain.} & \makecell[l]{\dots 4,433 were alive and not withdrawn at an\\\hll{LTFU} participating center.} & 1.1$\pm$0.4 & 16.6\% \\ \midrule
        \quad  General Domain & Abbreviations that belong to the general domain. & \text{\dots as low risk of bias (95\% \hll{CI} 0.37 to 1.53).} & 1.0$\pm$0.2 & 4.2\% \\

\bottomrule
    \end{tabular}
    }
    \vspace{-5pt}
    \caption{A taxonomy ($\mathcal{I}$) of complex textual spans in the medical domain with examples highlighted by a red background. The ``Medical Jargon'' and "Abbreviation" rows are based on the aggregation of sub-categories. }
    \vspace{-10pt}
    \label{tab:my_label}
    
\end{table*}

\subsection{Data Collection and Preprocessing}
\label{section:data-collection}

Different from prior work \cite{arase-etal-2022-cefr,naous2023towards}, our study consists of sentences from complete complex-simple article pairs, enabling a deeper analysis of how professional editors simplify medical documents.
The 15 resources that we considered include (1) the abstract sections and plain-language summaries from scientific papers, such as the National Institute for Health and Care Research (NIHR) and Cochrane Review of ``the highest standard in evidence-based healthcare'',\footnote{\url{https://www.cochranelibrary.com/}} for which we use the aligned article pairs released from prior studies \cite{devaraj-etal-2021-paragraph,goldsack-etal-2022-making, guo2022cells}; and (2) segment and paragraph pairs in the parallel versions of medical references from trusted online platforms, such as  Merck Manuals\footnote{\url{https://www.merckmanuals.com/}} and medical-related Wikipedia articles we extracted. A detailed description of each resource and pre-processing steps is provided in Appendix \ref{appendix:intro-resources}.

\paragraph{Target Audience.} To ensure our study reflects the background of a broader audience, our study mainly targets people who have completed high school or are entering college, and our dataset is annotated by college students without medical backgrounds using a six-point Likert scale.

\subsection{Sentence-level Readability Annotation}
\label{sec:sentence-readability}

To collect ground-truth judgments, we hire three university students with prior linguistic annotation experience to annotate the readability ratings for 4,520 sentences. We utilize the ``rank-and-rate'' interface \cite{naous2023towards} and the CEFR scale \cite{arase-etal-2022-cefr}, with several improvements.

\paragraph{Annotation Guidelines.} Following prior work \cite{arase-etal-2022-cefr}, we adopt the Common European Framework of Reference for Languages (CEFR) to annotate the sentence readability. CEFR standards were originally created for language learners. Because the scale is essentially a six-point Likert scale, we believe the findings would be mostly generalizable to a broader audience, including native speakers. Another reason for using the CEFR scale is to make our work comparable to the existing work and datasets which were created using the CEFR standards.

\paragraph{CEFR Scale.} CEFR is the most widely used international criteria to define learners’ language proficiency, assessing language skills on a 6-level scale with detailed guidelines,\footnote{\url{https://tinyurl.com/CEFR-Standard/}} from beginners (A1) to advanced mastery (C2), which are denoted as level 1 (easiest) to level 6 (hardest) in our interface. Following prior work \cite{arase-etal-2022-cefr, naous2023towards}, a sentence's readability is determined based on the CEFR level, at which an individual can understand the sentence without assistance. As medical texts naturally concentrate on the harder-to-understand side, we introduce the use of ``+'' and ``-'' signs to differentiate the nuance in readability, e.g., ``3+'' and ``3-'', in addition to each integer level. They are treated as 3.3 and 2.7 when converting to the numeric scores.

\paragraph{Rank-and-Rate Framework.} Six sentences are shown together to an annotator, who is instructed to rank them from most to least readable first, then rate each sentence using the 6-point CEFR standard. The interface is shown in Appendix \ref{appendix:annotation-interface}. Compared to rating each sentence individually, this method enables annotators to compare and contrast sentences within each set, leading to higher annotator agreement \cite{maddela-etal-2023-lens} and a more engaging user experience \cite{naous2023towards}.

\paragraph{Quality Control.}  For each medical sentence we annotate for the \medcsi~ corpus, we sample another (mostly non-medical) sentence with comparable length from the existing \textsc{ReadMe++} dataset \cite{naous2023towards} as a ``control''. Therefore, each set of sentences shown to the annotator consists of three medical sentences and three control sentences whose ratings are known. Annotators are asked to spend at least three minutes on every set, and their annotation quality is monitored through the use of control sentences. The  1,924 sentences in the dev and test sets are double annotated, and the scores are merged by average. The inter-annotator agreement is 0.742 measured by Krippendorff’s alpha \cite{krippendorff2011computing}. On the control sentences, our annotation achieves a Pearson correlation of 0.771 with the original ratings from \textsc{ReadMe++}.

\renewcommand{\arraystretch}{1.1}
\begin{table}[t!]
    \centering
    \small
    \vspace{-10pt}
    \resizebox{\linewidth}{!}{
    
    \begin{tabular}{p{2.35in} | c}
    \toprule
      \textbf{Feature}  & \textbf{Corr.} \\
\midrule 
Number of unique sophisticated lexical words$^\dagger$  & 0.645 \\
Corrected type-token-ratio (CTTR) & 0.627 \\
Number of syllables & 0.589 \\
Max age-of-acquisition (AoA) of words \shortcite{kuperman2012age} & 0.576 \\
Number of unique words & 0.574 \\
Number of words & 0.532 \\
Average number of characters per token & 0.524 \\
Corrected noun variation & 0.513 \\
The maximum dependency tree depth & 0.437 \\
Cumulative Zipf score for all words \shortcite{brysbaert2012adding} & 0.425 \\
     \bottomrule
         
    \end{tabular}
    }
    \vspace{-3pt}
    \caption{Top representative linguistic features and their Pearson correlation with readability. $^\dagger$Sophisticated lexical words \cite{lu2012relationship} are nouns, non-auxiliary verbs, adjectives, and certain adverbs that are not in the 2,000 most frequent lemmatized tokens in the \texttt{American National Corpus} (ANC).  More features and more implementation details are provided in the Appendix \ref{table:appendix-all-features}.  }
    \vspace{-15pt}
    \label{tab:feature_correlation_lexical}
\end{table}

\subsection{Fine-trained Complex Span Annotation}
\label{sec:jargon-annotation}
 We propose a new taxonomy to comprehensively capture 7 different categories of complex spans that appeared in the medical texts, as shown in Table \ref{tab:my_label}. The complete annotation guideline with more examples is provided in Appendix \ref{section:span-annotation-guideline}.

\paragraph{``Google-Hard'' Jargon.} In pilot study, we find that some medical terms, such as ``Tiotropium bromide'' (\textit{a drug}) and ``Plasmodium'' (\textit{an insect}), can be grasped after a quick Google search, although they are outside the vocabulary of many people. Some other phrases, such as ``anti-tumour necrosis factor failure'' and ``processive nucleases'', will require extensive research before a layperson can possibly (or still not) understand them, even though some of them contain short or common words. This seemingly minor distinction can have great implications in developing technological advances for medical text simplification and health literacy, motivating us to propose a novel category ``Google-Hard'' for medical jargon, which is separate from jargon that is ``Google-Easy'' or ``Name-Entity''.  In total, our dataset captures 698 Google-Hard medical jargon and 5,251 Google-Easy ones.

\renewcommand{\arraystretch}{1.12}
\begin{table}[t!]
    \centering
    \small
    \vspace{-10pt}
    
    \begin{tabular}{l|ccc}
    \toprule
      \textbf{Type}  & \textbf{\#Spans} & \textbf{\#Tokens} & \textbf{\%Tokens} \\
\midrule

Medical Jargon & 0.644 & 0.591 & 0.445 \\
Abbreviation  & 0.259 & 0.254 & 0.134\\
General Complex & 0.112 & 0.09 & 0.001  \\
Multi-sense  & 0.058 & 0.059 & 0.035  \\ 

\midrule
All Categories & 0.656 & 0.617 & 0.584 \\ 
     \bottomrule
         
    \end{tabular}

    \vspace{-3pt}
    \caption{The impact of 15 features related to complex spans, measured by the Pearson correlation with ground-truth sentence readability on the \medcsi~dataset.}
    \vspace{-15pt}
    \label{tab:feature_correlation}
\end{table}

\paragraph{Annotation Agreement.}

After receiving a two-hour training session, two of our in-hour annotators independently annotate each of the 4,520 sentences using a web-based annotation tool, BRAT \cite{stenetorp-etal-2012-brat}. The annotation interface is provided in Appendix \ref{section:span-annotation}. An adjudicator then further inspects the annotation and discusses any significant disagreements with the annotators. The inter-annotator agreement is 0.631 before adjudication, measured by token-level Cohen’s Kappa \cite{cohen1960coefficient}.

\section{Key Findings}
\label{sec:key-findings}
Enabled by our \medcsi~corpus, we first analyze the sentence readability measurements for medical texts (\S \ref{sec:why-medical-sentences-are-hard} and \S \ref{section:variance-of-readability}), then dive into medical jargon of different complexities (\S \ref{sec:what-makes-a-jargon-easy} and \S \ref{section:how-jargon-handle}).

\subsection{Why Medical Texts are Hard-to-Read?}
\label{sec:why-medical-sentences-are-hard}

 \begin{figure}[hpt!]
\centering
\vspace{-10pt}
\begin{subfigure}{0.505\linewidth}
  \centering
  \includegraphics[width=\linewidth]{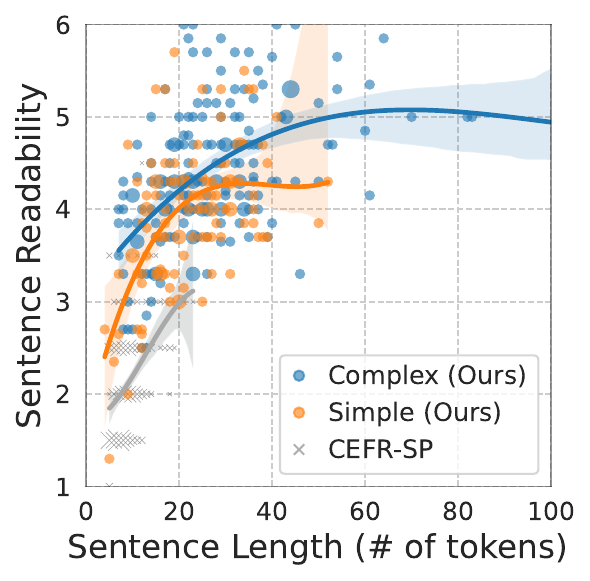}
  \label{fig:sub1}
\end{subfigure}%
\begin{subfigure}{0.495\linewidth}
  \centering
  \includegraphics[width=\linewidth]{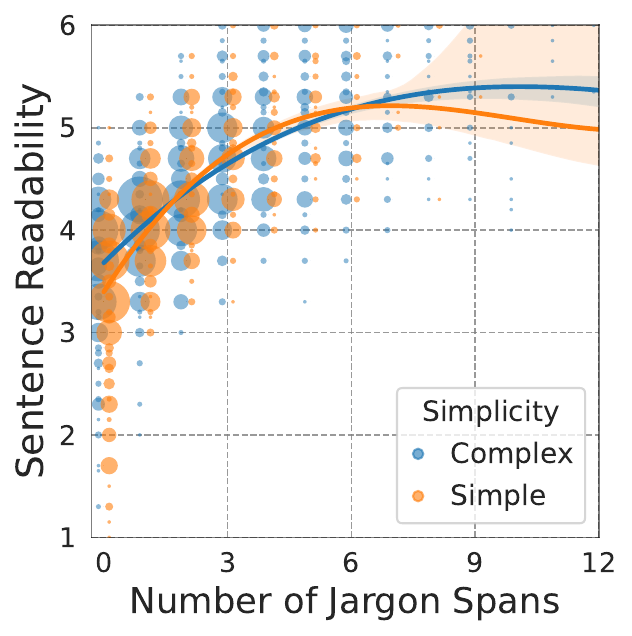}
  \label{fig:sub2}
\end{subfigure}
\vspace{-32pt}
\caption{\textit{Left}: Readability of sentences with different lengths. Compared to the CEFR-SP dataset \cite{arase-etal-2022-cefr}, our corpus contains much longer sentences. \textit{Right}: Readability of sentences with different numbers of jargon. The circle's radius reflects the number of overlapping points at each coordinate. We slightly shifted the points horizontally ($\pm$0.1) for better visualization. }
\vspace{-15pt}
\label{fig:two_distribution}
\end{figure}

The readability of a sentence can be impacted by a mixture of factors, including sentence length, grammatical complexity, word choice, etc. We extract 650 linguistic features from each sentence and measure their correlation with ground-truth readability. 15 additional features are designed to quantify the influence of complex spans. Based on our qualitative analysis, we found that complex spans, such as medical jargon, have a more profound impact on readability compared to other linguistic aspects.

\label{section:jargon-drive-up}

\paragraph{Impact of linguistic features.}  For each sentence, 650 linguistic features are extracted, including syntax and semantics features, quantitative and corpus linguistics features, in addition to psycho-linguistic features \cite{vajjala2016readability}, such as the age of acquisition (AoA) released by \citet{kuperman2012age}, and concreteness, meaningfulness, and imageability extracted from the MRC psycholinguistic database \cite{wilson1988mrc}. These features are extracted using a combination of toolkits,  each of which covers a different subset of features, including \texttt{LFTK} \cite{lee-lee-2023-lftk}, \texttt{LingFeat}, \texttt{Profiling–UD} \cite{brunato2020profiling}, \texttt{Lexical Complexity Analyzer} \cite{lu2012relationship}, and \texttt{L2 Syntactic Complexity Analyzer} \cite{lu2010automatic}.  We select and present top-10 representative features in Table \ref{tab:feature_correlation_lexical}, and provide a more complete list of the top-50 influential features in Appendix \ref{table:appendix-all-features} with more detailed definition of each feature.  We found that resource-based methods, such as the count of ``sophisticated lexical words'' \cite{lu2012relationship} and Zipf score \cite{powers-1998-applications}, are very useful. Length-related features are also informative.

\paragraph{Impact of Complex Spans.} 
Based on our pilot study and feedback from annotators, we observed that the specialized terminology, while allowing for precise and concise communication among experts,  significantly affects the difficulty level of texts in specialized domains. With our fine-grained span-level annotations (\S \ref{sec:jargon-annotation}), we can directly measure the effects that each type of complex words and jargon have on readability. 
Specifically, we design three features ``number-of-jargon-spans'', ``number-of-jargon-tokens'', and ``percentage-of-jargon-tokens'' for complex span in each category: \textit{medical jargon}, \textit{abbreviation}, \textit{general complex terms}, and \textit{multi-sense words}. We then compute their correlation with the sentence-level readability ratings. As shown in Table \ref{tab:feature_correlation}, we find that medical jargon significantly affects readability, and abbreviations follow in influence.

Figure \ref{fig:two_distribution} plots the relationship between readability and both the number of jargon spans (\textit{right}) and sentence length (\textit{left}), where sentences are split by whitespace.  On the left, we notice that the lines representing ``complex'' and ``simple'' sentences begin to diverge as sentence length exceeds 20 tokens, suggesting that factors beyond length affect the readability. In contrast, a stronger overall correlation between the number of jargon spans and readability is observed in the right figure.

\begin{figure}[t!]
    \centering
    \vspace{-10pt}
    \includegraphics[width=1\linewidth]{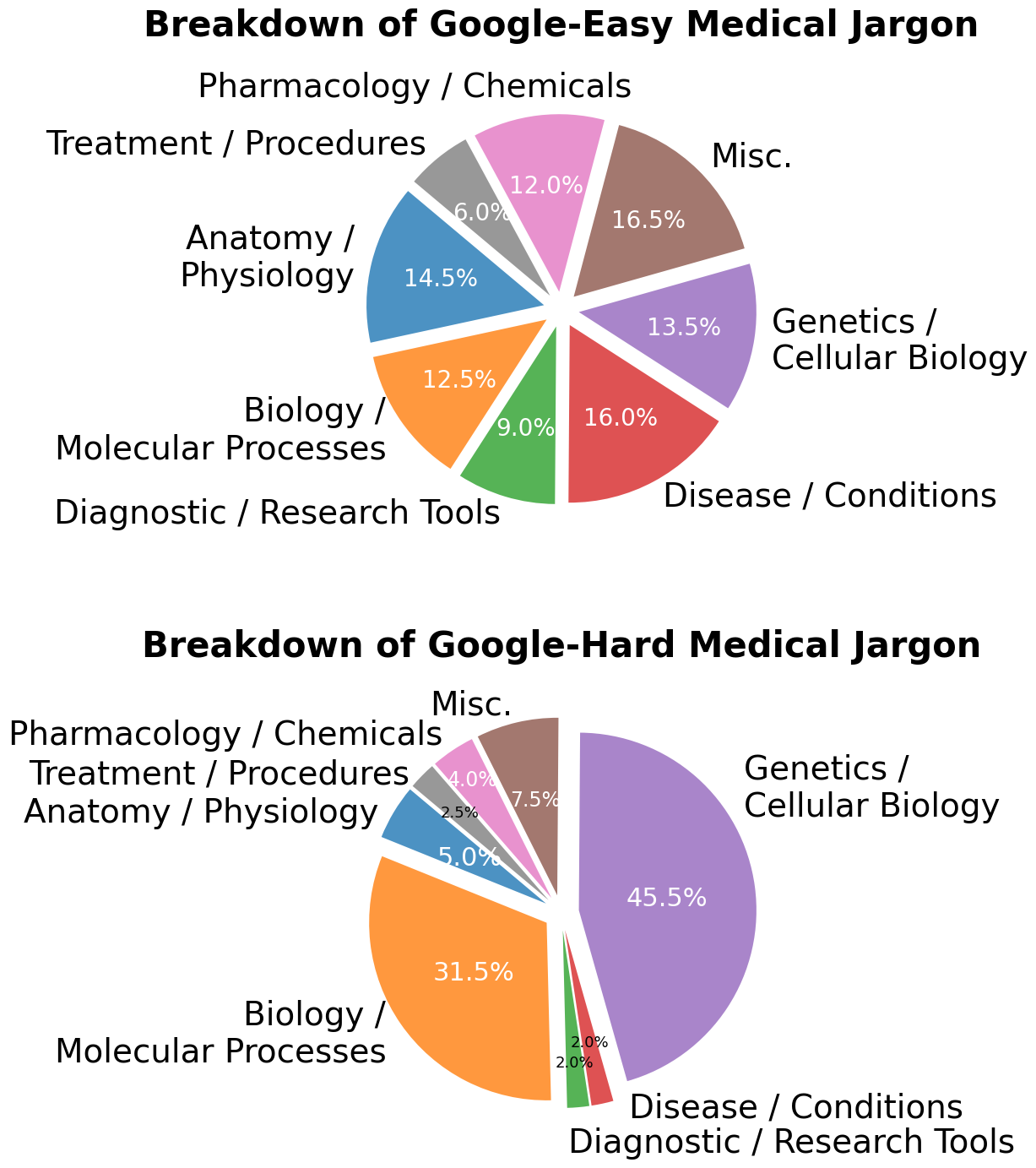}
    \vspace{-17pt}
    \caption{Breakdown of Google-Easy and Google-Hard jargon into different medical domains based on our manual analysis of 400 randomly sampled jargon.}
    \vspace{-15pt}
    \label{figure:breakdown-jargon-pie-chart}
\end{figure}

\subsection{What Makes a Jargon Easy (or Hard)?}

\label{sec:what-makes-a-jargon-easy}
Based on the feedback from annotators, we identify two major factors that influence the perceived difficulty of medical jargon, as listed below:

\paragraph{Inherent Complexity of Topics.} To analyze the perceived difficulty of medical jargon from different domains, we randomly sample 200 Google-Easy and 200 Google-Hard medical jargon, and manually analyze their topics.  The results are presented in Figure \ref{figure:breakdown-jargon-pie-chart}. Google-Easy terms are more diversified across different topics, while Google-Hard terms mainly fall under \textit{Genetics / Cellular Biology} and \textit{Biology / Molecular Processes}. This suggests that jargon associated with genetics or molecular procedures tends to be more challenging to read, possibly due to the specialized knowledge required to interpret them.

\renewcommand{\arraystretch}{1.3}

\setlength\tabcolsep{3pt} %
\begin{table*}[pht!]
    \centering
    \small
\vspace{-13pt}
\resizebox{\linewidth}{!}{
    \begin{tabular}{ l |  c |   cccc |  cccc}
    \toprule

                \textbf{Sources}  & \textbf{Length} & \makecell[c]{\textbf{FKGL}\\(\citeauthor{kincaid1975derivation})} & \makecell[c]{\textbf{ARI}\\(\citeauthor{smith1967automated})}  & \makecell[c]{\textbf{SMOG}\\(\citeauthor{mc1969smog})}     & \makecell[c]{\textbf{RSRS}\\(\citeauthor{martinc2021supervised})}   & \makecell[c]{\textbf{FKGL-Jar}\\(Ours)}  & \makecell[c]{\textbf{ARI-Jar}\\(Ours)} & \makecell[c]{\textbf{SMOG-Jar}\\(Ours)}  & \makecell[c]{\textbf{RSRS-Jar}\\(Ours)} \\

        \midrule

Cochrane & \gradecolor{62} 0.628 & \gradecolor{74} 0.743 & \gradecolor{68} 0.689 & \gradecolor{74} 0.749 & \gradecolor{82} 0.826 & \gradecolor{71} 0.717  & \gradecolor{71} 0.719  & \gradecolor{72} 0.726  & \gradecolor{72} 0.721 \\
PNAS & \gradecolor{55} 0.554 & \gradecolor{48} 0.480 & \gradecolor{44} 0.441 & \gradecolor{61} 0.615 & \gradecolor{59} 0.594 & \gradecolor{66} 0.660  & \gradecolor{65} 0.650  & \gradecolor{68} 0.685  & \gradecolor{65} 0.657 \\
NIHR Series & \gradecolor{52} 0.529 & \gradecolor{48} 0.482 & \gradecolor{45} 0.455 & \gradecolor{66} 0.661 & \gradecolor{65} 0.659 & \gradecolor{57} 0.577  & \gradecolor{58} 0.583  & \gradecolor{63} 0.632  & \gradecolor{61} 0.616 \\
eLife & \gradecolor{50} 0.505 & \gradecolor{19} 0.196 & \gradecolor{24} 0.244 & \gradecolor{37} 0.371 & \gradecolor{46} 0.467 & \gradecolor{64} 0.644  & \gradecolor{63} 0.638  & \gradecolor{69} 0.690  & \gradecolor{73} 0.733 \\
PLOS Series & \gradecolor{43} 0.436 & \gradecolor{41} 0.414 & \gradecolor{41} 0.413 & \gradecolor{44} 0.446 & \gradecolor{61} 0.613 & \gradecolor{71} 0.716  & \gradecolor{71} 0.717  & \gradecolor{70} 0.704  & \gradecolor{70} 0.707 \\
Wiki & \gradecolor{35} 0.352 & \gradecolor{40} 0.400 & \gradecolor{36} 0.368 & \gradecolor{47} 0.471 & \gradecolor{67} 0.670  & \gradecolor{67} 0.677  & \gradecolor{68} 0.681  & \gradecolor{78} 0.785  & \gradecolor{70} 0.703 \\
MSD & \gradecolor{25} 0.259 & \gradecolor{61} 0.618 & \gradecolor{57} 0.576 & \gradecolor{60} 0.604 & \gradecolor{69} 0.694 & \gradecolor{83} 0.836  & \gradecolor{83} 0.835  & \gradecolor{80} 0.805  & \gradecolor{85} 0.859 \\
\midrule
\textbf{Mean ± Std} & \gradecolor{46} 0.466 $\pm$ 0.127 & \gradecolor{47} 0.476 $\pm$ 0.173 & \gradecolor{45} 0.455 $\pm$ 0.143 & \gradecolor{56} 0.56 $\pm$ 0.134 & \gradecolor{64} 0.646 $\pm$ 0.109 & \gradecolor{69} 0.690 $\pm$  0.080    & \gradecolor{68} 0.689 $\pm$  0.080    & \gradecolor{71} 0.718 $\pm$  0.060    & \gradecolor{71} 0.714 $\pm$  0.076   \\
        
         \bottomrule
    \end{tabular}

}
    \vspace{-6pt}
    \caption{Pearson correlation  ($\uparrow$) between human ground-truth readability and each \textbf{unsupervised} readability metric. NIHR and PLOS are aggregations of 5 sources for each. All correlations are statistically significant. ``\textbf{-Jar}'' denotes adding a ``number-of-jargon'' feature into existing readability formula (more details in \S \ref{section:add-jar-method}). Our proposed method significantly improves the correlation over existing metrics, as demonstrated by the average correlation.}

    \vspace{-13pt}
    \label{tab:main_table_readability_unsupervised}
\end{table*}

\renewcommand{\arraystretch}{1.1}
\begin{table}[t!]
    \centering
    \small
    \begin{tabular}{l | c c }
    \toprule
        \textbf{Operation} &  \textbf{Google-Easy} & \textbf{Google-Hard} \\
\midrule
    \multicolumn{3}{l}{\textbf{\textit{Knowledge Panel}}} \\
    \midrule
        Covered & 45.6\% & 10.3\%   \\
        Explained by Figure & 13.6\% & 4.6\%   \\
        \midrule
    \multicolumn{3}{l}{\textbf{\textit{Feature Snippets}}} \\
    \midrule
        Covered  & 55.3\% & 21.2\%   \\
        Highlighted Text  & 52.4\% & 18.5\%  \\
        Explained by Figure & 22.8\% & 3.6\%   \\
    \bottomrule
    \end{tabular}
    \vspace{-4pt}
    \caption{The percentage of explanatory content provided by Google. An annotated screenshot of the webpage  is provided in Figure \ref{fig:annotated-google-screenshot} in Appendix \ref{appendix:annotated-google} to  visually demonstrates \textit{``Knowledge Panel''}  and \textit{``Feature Snippets''},}
    \vspace{-13pt}
    \label{table:how-jargon-display-google}
\end{table}

\paragraph{Variance in the Explanation.}  We also observed that the accessibility of medical jargon is greatly improved when search engines offer explanations or visual aids in their results. Search engines may provide the explanation of a medical term in two places: (1) the feature snippets in the answer box; and (2) the knowledge panel, which is powered by a knowledge graph. An annotated screenshot of the search results is provided in Figure \ref{fig:annotated-google-screenshot} in Appendix \ref{appendix:annotated-google} to demonstrate each element visually.  By parsing the Google search results for 2,731 unique Google-Easy and 504 Google-Hard medical jargon from our corpus, we quantified the existence of these explanations in Table \ref{table:how-jargon-display-google}.  The Google-Easy jargon is more frequently accompanied by explanatory content compared to the Google-Hard category. The use of visual aids also follows a similar pattern; Google-Easy terms are much more likely to be explained by figures compared to Google-Hard ones.

\renewcommand{\arraystretch}{1.2}
\begin{table}[pht!]
    \centering
    \small
    \begin{tabular}{l | c c }
    \toprule
        \textbf{Operation} &  \textbf{Google-Easy} & \textbf{Google-Hard} \\
    \midrule
        Kept & 22\% & 13\%  ($\downarrow 9\%$) \\
        Deleted & 56\% & 52\%  ($\downarrow 4\%$) \\
        Rephrased  & 3\% & 10\%   ($\uparrow 7\%$) \\

        Kept + Explained  & 8\% & 8\%  ($ -$) \\
        Del.+ Explained & 11\% & 17\%  ($\uparrow 6\%$) \\

    \bottomrule
    \end{tabular}
    \vspace{-3pt}
    \caption{The distribution of operations to 200 medical jargon (100 in each type), based on our manual analysis.}
    \vspace{-15pt}
    \label{table:how-jargon-handled}
\end{table}

\subsection{How Professional Editors Simplify the Medical Jargon?}  

\label{section:how-jargon-handle}

To study how jargon are handled during the manual simplification process, we randomly sample 200 jargon and manually analyze the operation applied to them. The results are presented in Table \ref{table:how-jargon-handled}. We find that the majority part of jargon in both categories got deleted. Compared to Google-Easy, ``Google-Hard'' jargon got copied less, and are being rephrased and explained more often. This findings indicate that trained editors adopt different strategies to handle jargon with different complexities.

\subsection{Readability Significantly Varies Across Existing Medical Simplification Corpora}
\label{section:variance-of-readability}

To better understand the quality of medical text simplification corpora, in Figure \ref{fig:readability-jargon-distribution}, we plot the distribution of sentence readability and numbers of jargon per sentence across 15 different resources. Within each source, the simplified texts are rated as easier to understand than their complex counterparts, though the extent varies. However, when compared across the board, simplified texts from some sources can be even more challenging to read than the complex texts from other sources, suggesting that not all plain texts are equally simple.  In addition, some resources, such as ``PLOS pathogens'', are especially difficult for laypersons without domain-specific knowledge to understand. The current research practice in medical text simplification often treat all data uniformly, such as concatenating all available corpora into one giant training set. However, we argue for a more cautious approach. For some resources, the ``simplified'' version remains quite complex, and the topics may not be directly relevant to laypersons.  Therefore, the decision to include a corpus or not should be made after considering the intended audiences' desired readability level and their use cases.

\section{Medical Readability Prediction}

In this section, we present a comprehensive evaluation of state-of-the-art readability metrics for medical texts (\S \ref{section:evaluating-existing-readability}), and design a simple yet effective method to further improve them (\S \ref{section:add-jar-method}).

\subsection{Evaluating Existing Readability Metrics}
Enabled by our annotated corpus, we first evaluate
commonly used sentence readability metrics.

\label{section:evaluating-existing-readability}

\paragraph{Unsupervised Methods.} The Pearson correlations between ground-truth readability and each unsupervised metric are presented in the left half of Table \ref{tab:main_table_readability_unsupervised}. The metrics we considered include FKGL \cite{kincaid1975derivation}, ARI \cite{smith1967automated}, SMOG \cite{mc1969smog}, and RSRS \cite{martinc2021supervised}, and their detailed formulations are provided in Appendix \ref{section:formula-of-existing-readability}. We also add sentence length as a baseline.  We find that the unsupervised methods generally do not perform very well. The language model-based RSRS score significantly outperforms the traditional feature-based metrics, among which SMOG performs best.

\paragraph{Supervised and Prompt-based Methods.}
The results are presented in Table \ref{tab:main_table_readability_supervised}. For supervised methods, we fine-tune language models on our dataset and existing corpora \cite{naous2023towards, arase-etal-2022-cefr, brunato-etal-2018-sentence} to predict the sentence readability. We also evaluate the performance of in-context learning by prompting large language models such as GPT-4 and Llama-3\footnote{More specifically, we used \texttt{gpt-4-0613} and \texttt{Llama-3.1-8B-Instruct} in the experiments.} \cite{llama3modelcard} using 5-shot. 
The prompts are constructed following \citet{naous2023towards}. More details and the full prompt template are in Appendix \ref{appendix:readability-prompt-template}. 
 We find that prompt-based methods achieve competitive results, e.g., GPT-4 outperforms the strongest unsupervised metric RSRS, although they still fall behind supervised methods.

\renewcommand{\arraystretch}{1.2}
\setlength\tabcolsep{3pt} %
\begin{table*}[pht!]
    \centering
    \small
\vspace{-4pt}
\resizebox{\linewidth}{!}{
    \begin{tabular}{ l |  c  c   |cccc | c ccc}
    \toprule

    \multirow{2}{*}{\textbf{Sources}}  & \multicolumn{2}{c|}{\textbf{5-shots}} & \multicolumn{4}{c|}{\raisebox{0.3mm}{\begin{minipage}{.033\textwidth}
      \includegraphics[height=4.5mm]{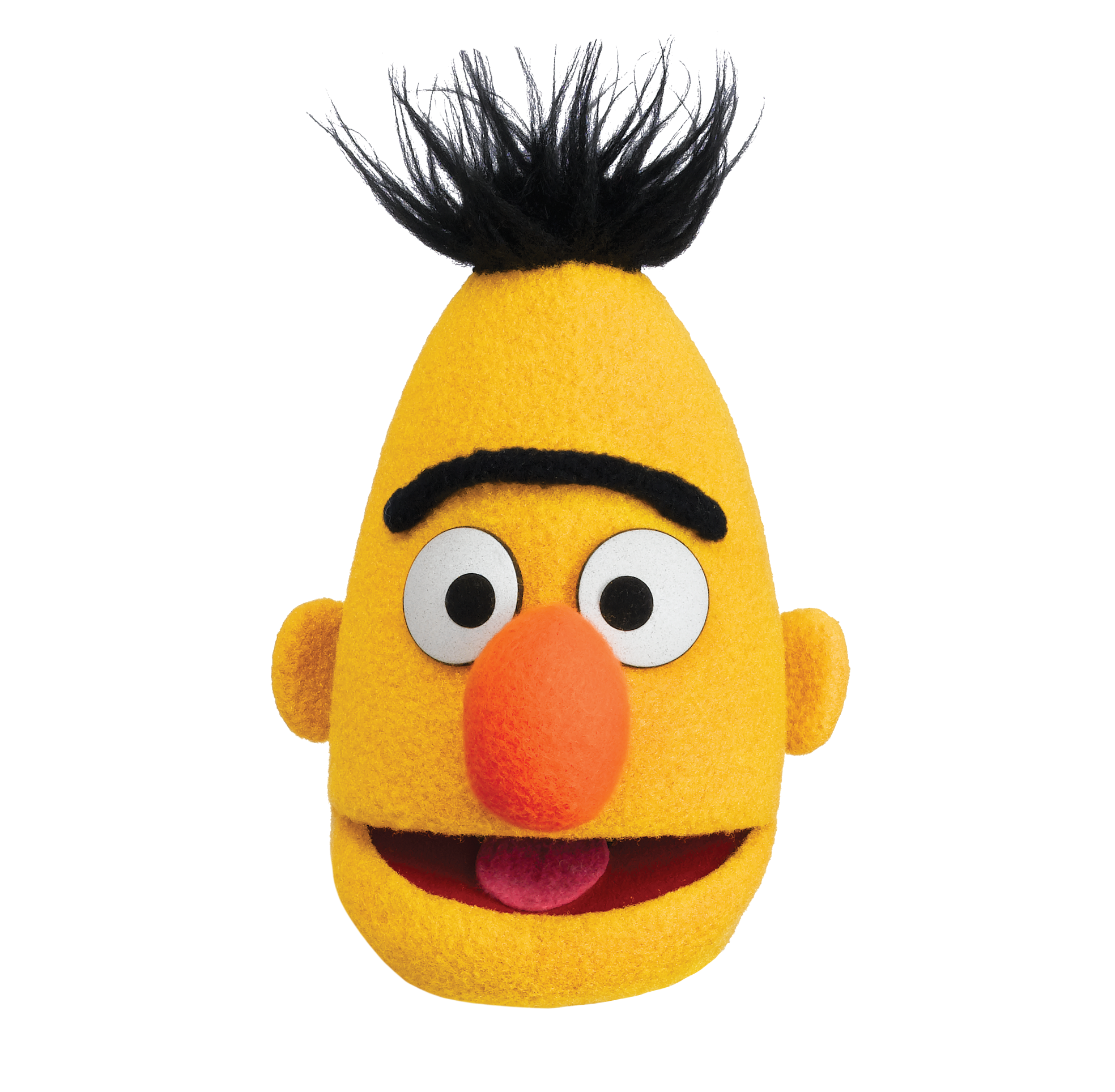}
    \end{minipage}}\textbf{Trained on Each Corpus}}  & \multicolumn{4}{c}{\textbf{The Trained}\raisebox{0.3mm}{\begin{minipage}{.033\textwidth}
      \includegraphics[height=4.5mm]{figs/Bert-Head.png}
    \end{minipage}}\textbf{+ an Jargon Term}} \\ \cmidrule{2-11}

& \makecell[c]{GPT-4\\(\citeauthor{achiam2023gpt})} &  \makecell[c]{Llama 3-8b\\(\citeauthor{llama3modelcard})} & \makecell[c]{ReadMe++\\(\citeauthor{naous2023towards})}  & \makecell[c]{CEFR-SP\\(\citeauthor{arase-etal-2022-cefr})}      & \makecell[c]{CompDS\\(\citeauthor{brunato-etal-2018-sentence})} & \makecell[c]{\medcsi\\(Ours)} & \makecell[c]{ReadMe++$_\textbf{Jar}$\\(Ours)}  & \makecell[c]{CEFR-SP$_\textbf{Jar}$\\(Ours)}     & \makecell[c]{CompDS$_\textbf{Jar}$\\(Ours)}  & \makecell[c]{\medcsi$_\textbf{Jar}$\\(Ours)}  \\

        \midrule

Cochrane &   \gradecolor{90} 0.908 & \gradecolor{66} 0.665 & \gradecolor{85} 0.858 & \gradecolor{89} 0.899 & \gradecolor{87} 0.870 & \gradecolor{94} 0.947 & \gradecolor{84} 0.842  & \gradecolor{85} 0.850  & \gradecolor{78} 0.785  & \gradecolor{88} 0.882  \\
PNAS & \gradecolor{78} 0.780 & \gradecolor{52} 0.528 & \gradecolor{85} 0.852 & \gradecolor{82} 0.820 & \gradecolor{79} 0.791 & \gradecolor{87} 0.874 & \gradecolor{78} 0.780  & \gradecolor{82} 0.824  & \gradecolor{74} 0.744  & \gradecolor{87} 0.873  \\
 NIHR Series &  \gradecolor{71} 0.713 & \gradecolor{48} 0.485 & \gradecolor{82} 0.824 & \gradecolor{75} 0.753 & \gradecolor{70} 0.706 & \gradecolor{88} 0.885& \gradecolor{69} 0.697  & \gradecolor{68} 0.687  & \gradecolor{63} 0.634  & \gradecolor{70} 0.700  \\

eLife &  \gradecolor{53} 0.538 & \gradecolor{18} 0.188 & \gradecolor{59} 0.594 & \gradecolor{71} 0.715 & \gradecolor{60} 0.608 & \gradecolor{71} 0.712 & \gradecolor{81} 0.812  & \gradecolor{80} 0.802  & \gradecolor{77} 0.777  & \gradecolor{86} 0.861  \\

PLOS Series &  \gradecolor{67} 0.672 & \gradecolor{52} 0.520 & \gradecolor{68} 0.680 & \gradecolor{69} 0.691 & \gradecolor{63} 0.635 & \gradecolor{70} 0.702 & \gradecolor{78} 0.787  & \gradecolor{84} 0.843  & \gradecolor{74} 0.744  & \gradecolor{85} 0.850 \\

Wiki & \gradecolor{67} 0.670 & \gradecolor{44} 0.447 & \gradecolor{82} 0.824 & \gradecolor{70} 0.709 & \gradecolor{60} 0.607 & \gradecolor{84} 0.843 & \gradecolor{71} 0.712  & \gradecolor{61} 0.619  & \gradecolor{67} 0.673  & \gradecolor{70} 0.709 \\

MSD &   \gradecolor{76} 0.766 & \gradecolor{56} 0.562 & \gradecolor{78} 0.784 & \gradecolor{77} 0.778 & \gradecolor{75} 0.757 & \gradecolor{86} 0.867 & \gradecolor{91} 0.918  & \gradecolor{88} 0.880  & \gradecolor{86} 0.863  & \gradecolor{93} 0.937  \\

\midrule
\textbf{Mean ± Std} & \gradecolor{72} 0.721 $\pm$ 0.115 & \gradecolor{48} 0.485 $\pm$ 0.148 & \gradecolor{77} 0.774 $\pm$ 0.1 & \gradecolor{76} 0.766 $\pm$ 0.073 & \gradecolor{71} 0.711 $\pm$ 0.101 & \gradecolor{83} 0.833 $\pm$ 0.092 & \gradecolor{79} 0.793 $\pm$  0.076    & \gradecolor{78} 0.786 $\pm$  0.096    & \gradecolor{74} 0.746 $\pm$  0.075    & \gradecolor{83} 0.830 $\pm$  0.090    \\

         \bottomrule
    \end{tabular}

}
    \vspace{-5pt}
    \caption{Pearson correlation ($\uparrow$) between human ground-truth readability and each \textbf{prompting} and \textbf{supervised} readability metric. All numbers are averaged over five runs, and all correlations are statistically significant. \raisebox{0.3mm}{\begin{minipage}{.03\textwidth}
      \includegraphics[height=4.5mm]{figs/Bert-Head.png}
    \end{minipage}} denotes RoBERTa-large models. ``\textbf{-Jar}'' means adding a ``jargon'' term (more details in \S \ref{section:add-jar-method}). Prompt-based methods are competitive, while still outperformed by fine-tuned models in much smaller sizes.}
    \label{tab:main_table_readability_supervised}
    \vspace{-7pt}
\end{table*}

\subsection{Improving Readability Metrics with Jargon Identification}
\label{section:add-jar-method}
To incorporate the consideration of jargon into existing metrics, we add and tune a  weight $\alpha$ for the feature ``number-of-jargon'' as follows:

\setlength{\abovedisplayskip}{-8pt}
\setlength{\belowdisplayskip}{6pt}
\setlength{\abovedisplayshortskip}{-8pt}
\setlength{\belowdisplayshortskip}{6pt}

\[
\text{FKGL-Jar} = \text{FKGL} + \alpha \times \text{\#Jargon},
\]

\setlength{\abovedisplayskip}{12pt plus 3pt minus 9pt}
\setlength{\belowdisplayskip}{12pt plus 3pt minus 9pt}
\setlength{\abovedisplayshortskip}{0pt plus 3pt}
\setlength{\belowdisplayshortskip}{6.5pt plus 3.5pt minus 3pt}

\noindent where ``FKGL-Jar'' denotes adding jargon into the FKGL score, similarly for other metrics with a suffix ``-Jar''. The weight  $\alpha$ is chosen by grid search on the dev set using gold annotation for each metric. As RSRS scores are smaller than 1, we scale them by 100 before the parameter search.  The right sides in Table \ref{tab:main_table_readability_unsupervised} and \ref{tab:main_table_readability_supervised} report the performance of each unsupervised and supervised method on the test set, after adding our proposed term.
To reflect the real-world scenario, we use jargon predicted by our best-performing complex span identification model (more details in \S \ref{sec:cwi_experiment}), instead of the ground-truth annotation. The optimal weights ($\alpha$) we tuned for ``FKGL-Jar'', ``ARI-Jar'', ``SMOG-Jar'', and ``RSRS-Jar'' are 4.85, 6.43, 1.1, and 0.45, respectively.  We find that introducing a single term significantly improves the correlation with human judgments.

\begin{figure*}[pht!]
    \centering
    \includegraphics[width=\linewidth]{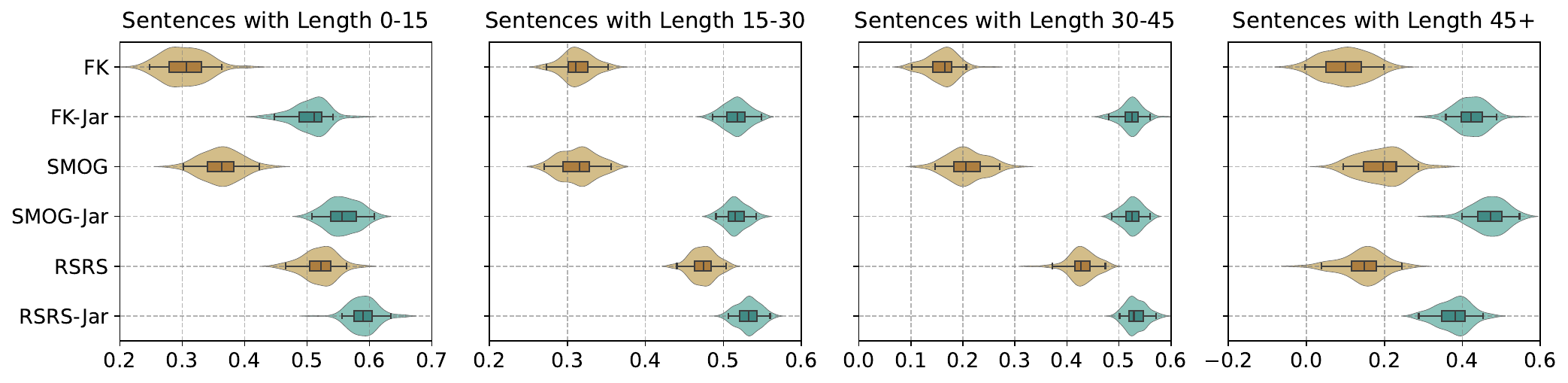}
    \vspace{-20pt}
    \caption{The 95\% confidence intervals for Kendall Tau-like correlation ($\uparrow$) between ground-truth readability annotation and predicted outputs from each automatic metric for sentences with different lengths, calculated by bootstrapping \cite{deutsch2021statistical}. In addition to a higher correlation with human judgments, incorporating jargon (``-Jar'') makes each metric more stable, as shown by the smaller intervals.}

    \vspace{-15pt}
    \label{fig:confidence_interval}
\end{figure*}

\paragraph{Length-Controlled Experiment.} To analyze the impact on sentences of varied lengths, in Figure \ref{fig:confidence_interval},  we present the 95\% confidence intervals for the Kendall Tau-like correlation \cite{noether1981kendall}  between the ground-truth readability and predictions from each metric \cite{maddela-etal-2023-lens}. We find the proposed ``-Jar'' term is advantageous for sentences at all lengths and is especially helpful for feature-based methods, such as SMOG. In addition, the incorporation of jargon makes the metrics more stable, as demonstrated by the narrower intervals.

\section{Fine-grained Complex Span Identification} 
\label{sec:cwi_experiment}

Based on our analysis in \S \ref{section:add-jar-method}, identifying complex spans in a sentence can help the judgment of its readability. It can also improve the performance of downstream text simplification system \cite{shardlow-2014-open}.
We formulate this task as a NER-style sequential labeling problem \cite{gooding-kochmar-2019-complex}, and utilize our annotated dataset to train and evaluate several models.

\paragraph{Data and Models.} The 4,520 sentences in our corpus is split into 2,587/784/1,140 for train, dev, and test sets. We mainly consider BERT/RoBERTa-based standard tagging models, initialized with different pre-trained embeddings. The implementation details are provided in Appendix \ref{appendix:cwi_iumplementation_details}.

\paragraph{Evaluation Metrics.} We consider two variants of F1 measurements:  (1) {entity-level partial match}, indicating the number of jargon, where the type of the predicted entity matches the gold entity and the predicted boundary overlaps with the gold span. We use the evaluation script released by \citet{tabassum-etal-2020-wnut}.\footnote{ \url{https://github.com/jeniyat/WNUT_2020_NER/tree/master/code/eval}} We also report the exact match performance at entity-level in the Appendix \ref{appendix:cwi_imore_performance}. (2) {token-level match}, measuring the number of jargon tokens.\ For each metric, we conduct evaluations at three levels of granularity: (1) fine-grained level with 7 categories, (2) associated 3 higher-level classes (i.e., medical / general+multisense / abbreviation), and (3) binary judgments between complex or non-complex text spans.

\paragraph{Results.} The evaluation results are presented in Table \ref{table:complex-span-identification}. All results are averaged over 5 runs with different random seeds. The fine-tuned RoBERTa-large model \cite{liu2019roberta} achieves 86.8 and 80.2 F1 for binary tasks at token- and entity levels. Using predictions from this model, we significantly improve existing readability metrics' correlation with human judgment (\S \ref{section:add-jar-method}). We find the domain-specific models at base size, such as PubMedBERT \cite{pubmedbert}, also achieve competitive performance. However, differentiating between the seven categories of complex spans remains challenging. 

\begin{table}[pht!]
\renewcommand{\arraystretch}{1.18}
\setlength\tabcolsep{2pt} %

\normalsize
\centering
\resizebox{\linewidth}{!}{
\begin{tabular}{@{\hspace{2.5pt}}l c@{\hspace{2.5pt}}c@{\hspace{3pt}}c@{\hspace{7pt}}  c@{\hspace{2.5pt}}c@{\hspace{3pt}}c@{\hspace{2pt}}}

\toprule

\multirow{2}{*}{\textbf{Models}} &  \multicolumn{3}{c}{\textbf{Token-Level}} &  \multicolumn{3}{c}{\textbf{Entity-Level}} \\

  &  Binary   & 3-Cls. & 7-Cate.   &  Binary   & 3-Cls. & 7-Cate.   \\ 

    \midrule

    \multicolumn{7}{l}{\textbf{\textit{Large-size Models}}}\\ [-0.1ex]
    \midrule

BERT \shortcite{devlin-etal-2019-bert} & 86.1 & 80.9 & 67.9 & 78.5 & 74.1 & 43.9 \\
RoBERTa \shortcite{liu2019roberta} & \textbf{86.8} & \textbf{82.3} & \textbf{68.6} & \textbf{80.2} & \textbf{75.9} & \textbf{67.9} \\
BioBERT \shortcite{lee2020biobert} & 85.3 & 80.7 & 67.0 & 78.4 & 72.6 & 64.9 \\
PubMedBERT \shortcite{pubmedbert} & 85.7 & \textbf{82.3} & \underline{68.3} & 79.0 & \underline{75.2} & 66.5 \\

    \midrule

    \multicolumn{7}{l}{\textbf{\textit{Base-size Models}}}\\ [-0.1ex]
    \midrule
    
BERT \shortcite{devlin-etal-2019-bert} & 85.4 & 80.4 & 66.3 & 77.0 & 72.5 & 63.3 \\
RoBERTa \shortcite{liu2019roberta} & \underline{86.2} & \underline{81.7} & 68.0 & \underline{79.7} & 75.2 & \underline{66.6} \\
BioBERT \shortcite{lee2020biobert} & 84.2 & 79.6 & 66.4 & 77.1 & 72.8 & 64.1 \\
PubMedBERT \shortcite{pubmedbert} & 85.2 & 81.2 & 67.7 & 78.5 & 74.8 & 66.3 \\

    \bottomrule
\end{tabular}
}
\vspace{-3pt}
\caption{\textbf{Micro F1} ($\uparrow$) of different systems for complex span identification on the {\medcsi} test set. The \textbf{best} and \underline{second-best} scores are highlighted. Models are trained with fine-grained labels in seven categories and evaluated at different granularity.}
\label{table:complex-span-identification}
\vspace{-13pt}
\end{table}

\begin{table}[tph!]
\vspace{+4pt}
\renewcommand{\arraystretch}{1.25}
\setlength\tabcolsep{1pt} %

\centering
\small
\resizebox{\linewidth}{!}{
\begin{tabular}{l@{\hspace{5pt}}cccc}

\toprule

\textbf{Training Corpus} & \textbf{Domain} & \textbf{\#Sent.} & \textbf{Token} & \textbf{Entity} \\
\midrule
SemEval2016 \shortcite{paetzold-specia-2016-semeval} & Wikipedia & 200 & 38.6 & 29.0 \\
CWIG3G2 \shortcite{yimam-etal-2017-cwig3g2} & News, Wiki & 1,988 & 46.4 & 28.7 \\
\midrule 
\medcsi~(Ours) & Medical Articles & 4,520 & 86.8 & 80.2 \\
    \bottomrule
\end{tabular}
}
\vspace{-3pt}
\caption{F1 on the test set of \medcsi~  for  models trained on different datasets. ``Entity'' and ``Token'' denote binary entity-/token-level performance. ``\#Sent'' is the number of unique sentences in the training set.}
\label{table:transfer-learning}
\vspace{-11pt}
\end{table}

\paragraph{Transfer Learning.}  We use two existing datasets \cite{paetzold-specia-2016-semeval, yimam-etal-2017-cwig3g2} to train RoBERTa-large \cite{liu2019roberta} models, and evaluated them on the test set of our  \medcsi.  Table \ref{table:transfer-learning} presents the performance for  binary complex span identification task, as existing corpora consist of binary labels, and SemEval2016 \cite{paetzold-specia-2016-semeval} only has complex word annotation. We find that both models trained using general domain data do not perform well in the medical field. This results demonstrate the necessity for our medical-focus dataset.

\section{Related Work}
\label{section:related-work}

\paragraph{Readability Measurement in Medical Domain.} Unsupervised metrics, such as FKGL \cite{kincaid1975derivation}, ARI \cite{smith1967automated}, SMOG \cite{mc1969smog}, and Coleman-Liau index \cite{coleman1975computer} have been widely adopted in existing research on the medical readability analysis, as they do not require training data \cite[\textit{inter alia}]{fu2016search, chhabra2018evaluation, xu2019easier, devaraj-etal-2021-paragraph, kruse2021readability, guo2022cells, kaya2022quality, hartnett2023readability}. However, their reliability has been questioned \cite{wilson2009readability, jindal2017assessing,devaraj2021paragraph}, as they mainly rely on the combination of shallow lexical features. Unsupervised RSRS score \cite{martinc2021supervised} utilizes the log probability of words from a pre-trained language model such as BERT \cite{devlin-etal-2019-bert}, while other supervised metrics rely on fine-tuning LLMs on the annotated corpora \cite{arase-etal-2022-cefr, naous2023towards}; however, previously, the performance of these methods on the medical texts were unclear. Enabled by our high-quality dataset, we benchmark existing state-of-the-art metrics in the medical domain (\S \ref{section:evaluating-existing-readability}), and also further improve their performances (\S \ref{section:add-jar-method}).

\paragraph{Complex Span Identification in Medical Domain.} \citet{kauchak2016moving} collects a dataset that consists of the difficulty for 275 words. CompLex 2.0 \cite{shardlow-etal-2020-complex} consists of complex spans rated on a 5-point Likert scale. However, it only covers spans with one or two tokens. MedJEx corpus \cite{kwon-etal-2022-medjex} consists of binary jargon annotation for sentences in the electronic health record (EHR) notes, whereas the dataset is licensed. Other work on complex word identification mainly focuses on general domains, such as news and Wikipedia, and other specialized domains, e.g., computer science. Due to space limits, we list them in Appendix \ref{sec:more-related-work}.  Our data is based on open-access medical resources and contains both sentence-level readability ratings and complex span annotation with a finer-grained  7-class categorization (\S \ref{section:dataset-construction}).

\section{Conclusion}
In this work, we present a systematic study for sentence readability in the medical domain, featuring a new annotated dataset and a data-driven study to answer \textit{“why medical sentences are so hard.”}. In the analysis, we quantitatively measure the impact of several key factors that contribute to the complexity of medical texts, such as the use of jargon, text length, and complex syntactic structures.  Future work could extend to the medical notes from clinical settings to better understand real-time communication challenges in healthcare. Additionally, leveraging our dataset that categorizes complex spans by difficulty and type, further research could develop personalized simplification tools to adapt content to the target audience, thereby improving patients' understanding of medical information.

\section*{Limitations}
Due to the reality that major scientific medical discoveries are mostly reported in English, our study primarily focuses on English-language medical texts. Future research could extend to medical resources in other languages. In addition, the focus of our work is to create readability datasets for general purposes following prior work. We did not study or distinguish the fine-grained differences and nuances between native speakers and non-native speakers \cite{yimam-etal-2017-cwig3g2}. 

The readability ratings of a sentence can be impacted by a mixture of factors, including sentence length, grammatical complexity, word difficulty, the annotator's educational background, the design and quality of annotation guidelines, as well as the target audience. We choose to use the CEFR standards, which is ``the most widely used international standard'' to access learners' language proficiency \cite{arase-etal-2022-cefr}. It has detailed guidelines in 34 languages\footnote{ \url{http://tinyurl.com/CEFR-Standard}}\textsuperscript{,}\footnote{\url{http://tinyurl.com/CEFR-34-languages}} and have been widely used in many prior research \cite[\textit{inter alia}]{boyd-etal-2014-merlin, rysova-etal-2016-automatic, francois-etal-2016-svalex, xia-etal-2016-text,tack-etal-2017-human, wilkens-etal-2018-sw4all, arase-etal-2022-cefr, naous2023towards}.

\section*{Ethics Statement}
During the data collection process, we hired undergrad students from the U.S. as in-house annotators. All annotators are compensated at \$18 per hour or by credit hours based on the university standards.

\section*{Acknowledgments}
The authors would like to thank Mithun Subhash, Jeongrok Yu, and Vishnu Suresh for their help in data annotation. This research is supported in part by the NSF CAREER Award IIS-2144493, NSF Award IIS-2112633, NIH Award R01LM014600, ODNI and IARPA via the HIATUS program (contract 2022-22072200004). The views and conclusions contained herein are those of the authors and should not be interpreted as necessarily representing the official policies, either expressed or implied, of NSF, NIH, ODNI, IARPA, or the U.S. Government. The U.S. Government is authorized to reproduce and distribute reprints for governmental purposes notwithstanding any copyright annotation therein.

\bibliography{anthology,custom}

\clearpage
\appendix

\section{Formulas of Readability Metrics}

\label{section:formula-of-existing-readability}
In this section, we list the formulas for four unsupervised readability metrics.

\paragraph{FKGL.} The Flesch-Kincaid Grade Level formula is a well-known readability test designed to indicate how difficult a text in English is to understand. It is calculated using the formula:
\[
\begin{aligned}
FKGL &= 0.39 \left( \frac{\text{total words}}{\text{total sentences}} \right) \\
&\quad+ 11.8 \left( \frac{\text{total syllables}}{\text{total words}} \right) \\
&\quad- 15.59
\end{aligned}
\]

\paragraph{ARI.} The Automated Readability Index (ARI) is another widely used readability metric that estimates the understandability of English text. It is formulated based on characters rather than syllables. The ARI formula is given by:

\[
\begin{aligned}
ARI &= 4.71 \left( \frac{\text{total characters}}{\text{total words}} \right) \\
&\quad+ 0.5 \left( \frac{\text{total words}}{\text{total sentences}} \right) \\
&\quad- 21.43
\end{aligned}
\]

\paragraph{SMOG.} The SMOG (Simple Measure of Gobbledygook) Index is a readability formula that measures the years of education needed to understand a piece of writing. SMOG is particularly useful for higher-level texts. The formula is as follows, where the polysyllables are calculated by counting the number of words in a text that have three or more syllables:
\[
\begin{aligned}
P &= \text{number of polysyllables} \\
S &= \text{number of sentences} \\
SMOG &= 1.0430 \sqrt{P \times \frac{30}{S}} + 3.1291
\end{aligned}
\]

\paragraph{RSRS.} The RSRS (Ranked Sentence Readability Score) leverages log probabilities from a neural language model and the sentence length feature. It's calculated through a weighted sum of individual word losses. Each word’s Negative Log Loss (WNLL) is sorted in ascending order and weighted by its rank.  The formula assigns higher weights to the out-of-vocabulary (OOV) words, by setting $\alpha = 2$ for all OOV words and 1 for others.  The formula for RSRS is:
$$RSRS = \frac{\sum_{i=1}^{S} [\sqrt{i}]^{\alpha} \cdot WNLL(i)}{S}$$
And WNLL can be calculated by:
$$WNLL = -(y_t \log y_p + (1 - y_t) \log(1 - y_p))$$

Here, $S$ is sentence length,  $y_p$ is the predicted distribution from the language model, and  $y_t$ is the empirical distribution, where 1 for  words that appear in the text, and 0 for all others.

\section{More Results on the Influence of Each Linguistic Feature}
\label{ref:appendix-more-features}
In this section, we provide more results on the influence of linguistic features, including syntax and semantics features, quantitative and corpus linguistics features, in addition to psycho-linguistic features \cite{vajjala2016readability}, such as the age of acquisition (AoA) released by \citet{kuperman2012age}, and concreteness, meaningfulness, and imageability extracted from the MRC psycholinguistic database \cite{wilson1988mrc}.

The features are extracted using a combination of toolkits,  each of which covers a different subset of features, including 220 features from the \texttt{LFTK} package \cite{lee-lee-2023-lftk}, 255 from the \texttt{LingFeat} \cite{lee-etal-2021-pushing}, 61 from \texttt{Text Characterization Toolkit (TCT)} \cite{simig-etal-2022-text}, 119 from \texttt{Profiling–UD} \cite{brunato2020profiling}, 33 from the \texttt{Lexical Complexity Analyzer (LCA)} \cite{lu2012relationship}  and 23 from the \texttt{L2 Syntactic Complexity Analyzer (L2SCA)} \cite{lu2010automatic}. The top 50 most influential features are presented in Table \ref{table:appendix-all-features} after skipping the duplicated and nearly equivalent ones, e.g., the \textit{typo-token-ratio} and \textit{root-type-token-ratio}. 

For each of the listed features, we look into the implementation details from the original toolkit and explain them in the "Implementation Details" column. To facilitate reproducibility, we also include the exact feature name used in the original code in the "Original Feature Name" column.

\clearpage

\noindent\begin{minipage}{\textwidth}

\resizebox{\linewidth}{!}{
\begin{tabular}{llcp{3.5in}}
\toprule
\textbf{Package} & \textbf{Original Feature Name} &  \textbf{\makecell[c]{Pearson\\Correlation}} & \textbf{Implementation Details in the Original Toolkit} 
\\
\midrule

LCA \shortcite{lu2012relationship} &   \texttt{len(slextypes.keys())}  & 0.6452 & Number of unique sophisticated lexical words, which are lexical words (i.e., nouns, non-auxiliary verbs, adjectives, and certain adverbs that provide substantive content in the text) and are also ``sophisticated'' (i.e., not in the list of 2,000 most frequent lemmatized tokens in the ANC\footnote{\url{https://anc.org/}} corpus).  \\
LCA \shortcite{lu2012relationship} &   \texttt{len(swordtypes.keys())}  & 0.6408 & Number of unique sophisticated words. ``Sophisticated'' is defined as not in the list of 2,000 most frequent lemmatized tokens in the \texttt{American National Corpus} (ANC) \\
LFTK \shortcite{lee-lee-2023-lftk} &   \texttt{corr\_ttr}  & 0.6271 & Corrected type-token-ratio (CTTR), which is calculated as $(\text{number-of-unique-tokens} / \sqrt{2 \times \text{number-of-all-tokens}})$, based on the lemmatized tokens.  \\
LFTK \shortcite{lee-lee-2023-lftk} &   \texttt{corr\_ttr\_no\_lem}  & 0.6158 & Corrected type-token-ratio (CTTR), which is calculated as $(\text{number-of-unique-tokens} / \sqrt{2 \times \text{number-of-all-tokens}})$, based on the  tokens without lemmatization. \\
LCA \shortcite{lu2012relationship} &   \texttt{slextokens}  & 0.6120 & Number of all sophisticated lexical words, which are lexical words (i.e., nouns, non-auxiliary verbs, adjectives, and certain adverbs that provide substantive content in the text) and are also ``sophisticated'' (i.e., not in the list of 2,000 most frequent lemmatized tokens in the ANC corpus). \\
LCA \shortcite{lu2012relationship} &   \texttt{swordtokens}  & 0.6083 & Number of all sophisticated words. ``Sophisticated'' is defined as not in the 2,000 most frequent lemmatized tokens in the \texttt{American National Corpus} (ANC) \\
LCA \shortcite{lu2012relationship} &   \texttt{ndwz}  & 0.6037 & Number of different words in the first Z words. Z is computed as the 20th percentile of word counts from a dataset, resulting in a value of 16 in our case. \\
LCA \shortcite{lu2012relationship} &   \texttt{ndwesz}  & 0.6024 &  Number of different words in expected random sequences of Z words over ten trials. Z is computed as the 20th percentile of word counts from a dataset, resulting in a value of 16 in our case. \\
LingFeat \shortcite{lee-etal-2021-pushing} &   \texttt{WRich20\_S}  & 0.6006 &  Semantic richness of a text, which is calculated by summing up the probabilities of 200 Wikipedia-extracted topics, each multiplied by its rank, indicating the text's variety and depth of topics. The 200 topics were extracted from the Wikipedia corpus using the Latent Dirichlet Allocation (LDA) method.  \\
LCA \shortcite{lu2012relationship} &   \texttt{len(lextypes.keys())}  & 0.5996 & Number of unique lexical words. Lexical words include nouns, non-auxiliary verbs, adjectives, and certain adverbs that provide substantive content in the text.  \\
LCA \shortcite{lu2012relationship} &   \texttt{ndwerz}  & 0.5961 &  Number of different words expected in random Z words over ten trials. Z is computed as the 20th percentile of word counts from a dataset, resulting in a value of 16 in our case.  \\
LFTK \shortcite{lee-lee-2023-lftk} &   \texttt{t\_syll}  & 0.5888 & Number of syllables. \\
LFTK \shortcite{lee-lee-2023-lftk} &   \texttt{t\_char}  & 0.5806 & Number of characters. \\
TCT \shortcite{simig-etal-2022-text} &   \texttt{WORD\_PROPERTY\_AOA\_MAX}  & 0.5758 & Max age-of-acquisition (AoA) of words. The AoA of each word is defined by \citet{kuperman2012age}. \\
LCA \shortcite{lu2012relationship} &   \texttt{lextokens}  & 0.5750 & Number of lexical words. Lexical words include nouns, non-auxiliary verbs, adjectives, and certain adverbs that provide substantive content in the text.\\

\bottomrule
\end{tabular}
}
 \\

\label{table:appendix-all-features}
\captionof{table}{Top 50 most influential linguistic features on readability assessment. }

\end{minipage}
\clearpage

\noindent\begin{minipage}{\textwidth}

\resizebox{\linewidth}{!}{
\begin{tabular}{llcp{3.5in}}
\toprule
\textbf{Package} & \textbf{Original Feature Name} &  \textbf{\makecell[c]{Pearson\\Correlation}} & \textbf{Implementation Details in the Original Toolkit} 
\\
\midrule

LFTK \shortcite{lee-lee-2023-lftk} &   \texttt{t\_uword}  & 0.5744 & Number of unique words. \\
LingFeat \shortcite{lee-etal-2021-pushing} &   \texttt{WTopc20\_S}  & 0.5686 & The count of distinct topics, out of 200 extracted from Wikipedia, that are significantly represented in a text, showing the breadth of topics it covers.\\
LFTK \shortcite{lee-lee-2023-lftk} &   \texttt{t\_syll2}  & 0.5607 & Number of words that have more than two syllables.\\
LingFeat \shortcite{lee-etal-2021-pushing} &   \texttt{BClar20\_S}  & 0.5598 & Semantic Clarity measured by averaging the differences between the primary topic's probability and that of each subsequent topic, reflecting how prominently a text focuses on its main topic, based on 200 topics extracted from the WeeBit Corpus.
 \\

LingFeat \shortcite{lee-etal-2021-pushing} &   \texttt{to\_AAKuW\_C}  & 0.5379 & Total age-of-acquisition (AoA) of words. The AoA of each word is defined by \citet{kuperman2012age}.   \\

TCT \shortcite{simig-etal-2022-text} &   \texttt{DESWC}  & 0.5323 & Number of words. \\
LingFeat \shortcite{lee-etal-2021-pushing} &   \texttt{BClar15\_S}  & 0.5294 & Semantic Clarity measured by averaging the differences between the primary topic's probability and that of each subsequent topic, reflecting how prominently a text focuses on its main topic, based on 150 topics extracted from the WeeBit Corpus. \\

LingFeat \shortcite{lee-etal-2021-pushing} &   \texttt{at\_Chara\_C}  & 0.5237 & Average number of characters per token. \\
LFTK \shortcite{lee-lee-2023-lftk} &   \texttt{corr\_noun\_var}  & 0.5127 & Corrected noun variation, which is computed as  $(\text{number-of-unique-nouns} / \sqrt{2 \times \text{number-of-all-nouns}})$ \\

LingFeat \shortcite{lee-etal-2021-pushing} &   \texttt{as\_AAKuW\_C}  & 0.5069 & Average age-of-acquisition (AoA) of words. The AoA of each word is defined by \citet{kuperman2012age}. \\
LFTK \shortcite{lee-lee-2023-lftk} &   \texttt{t\_bry}  & 0.5046 & Total age-of-acquisition (AoA) of words. The AoA of each word is defined by \citet{brysbaert2017test}.   \\
LFTK \shortcite{lee-lee-2023-lftk} &   \texttt{t\_syll3}  & 0.5044 & Number of words that have more than three syllables. \\
LingFeat \shortcite{lee-etal-2021-pushing} &   \texttt{WTopc15\_S}  & 0.4956 & The count of distinct topics, out of 150 extracted from Wikipedia, that are significantly represented in a text, showing the breadth of topics it covers.\\
LFTK \shortcite{lee-lee-2023-lftk} &   \texttt{corr\_adj\_var}  & 0.4764 & Corrected adjective variation, which is computed as $(\frac{\text{number-of-unique-adjectives}}  {\sqrt{2 \times \text{number-of-all-adjectives}}})$  \\

LFTK \shortcite{lee-lee-2023-lftk} &   \texttt{n\_unoun}  & 0.4694 & Number of unique nouns. \\
LingFeat \shortcite{lee-etal-2021-pushing} &   \texttt{at\_Sylla\_C}  & 0.4691 & Average number of syllables per token. \\
LFTK \shortcite{lee-lee-2023-lftk} &   \texttt{a\_bry\_ps}  & 0.4586 &  Average age-of-acquisition (AoA) of words. The AoA of each word is defined by \citet{brysbaert2017test}.\\
LFTK \shortcite{lee-lee-2023-lftk} &   \texttt{n\_noun}  & 0.4581 & Number of nouns. \\
LingFeat \shortcite{lee-etal-2021-pushing} &   \texttt{to\_FuncW\_C}  & 0.4515 & Number of function words, excluding words with POS tags of 'NOUN', 'VERB', 'NUM', 'ADJ', or 'ADV'.  \\
LFTK \shortcite{lee-lee-2023-lftk} &   \texttt{n\_adj}  & 0.4497 & Number of adjectives. \\
LFTK \shortcite{lee-lee-2023-lftk} &   \texttt{n\_uadj}  & 0.4483 & Number of unique adjectives. \\
Profiling–UD \shortcite{brunato-etal-2020-profiling} &   \texttt{avg\_max\_depth}  & 0.4371 & The maximum tree depths extracted from a sentence, which is calculated as the longest path (in terms of occurring dependency links) from the root of the dependency tree to some leaf. \\
LingFeat \shortcite{lee-etal-2021-pushing} &   \texttt{WNois20\_S}  & 0.4362 & Semantic noise, which quantifies the dispersion of a text's topics, reflecting how spread out its content is across different subjects. It is calculated by analyzing the text's topic probabilities on 200 topics extracted from through Latent Dirichlet Allocation (LDA). \\

LCA \shortcite{lu2012relationship} &   \texttt{ls1}  & 0.4255 & Lexical Sophistication-I, calculated as the ratio of sophisticated lexical tokens to the total number of lexical tokens.\\
\bottomrule
\end{tabular}
}

\label{table:appendix-all-features}
\captionof{table}{Top 50 most influential linguistic features on readability assessment (continue). }

\end{minipage}
\clearpage

\noindent\begin{minipage}{\textwidth}

\resizebox{\linewidth}{!}{
\begin{tabular}{llcp{3.5in}}
\toprule
\textbf{Package} & \textbf{Original Feature Name} &  \textbf{\makecell[c]{Pearson\\Correlation}} & \textbf{Implementation Details in the Original Toolkit} 
\\
\midrule

LFTK \shortcite{lee-lee-2023-lftk} &   \texttt{t\_subtlex\_us\_zipf}  & 0.4253 & Cumulative Zipf score for all words, based on frequency data from the SUBTLEX-US corpus \cite{brysbaert2012adding}. Zipf scores are a measure of word frequency, with higher scores indicating more common words. \\
LingFeat \shortcite{lee-etal-2021-pushing} &   \texttt{WTopc10\_S}  & 0.4242 &  The count of distinct topics, out of 100 extracted from Wikipedia, that are significantly represented in a text, showing the breadth of topics it covers.\\
Profiling–UD \shortcite{brunato-etal-2020-profiling} &   \texttt{avg\_links\_len}  & 0.4167 & Average number of words occurring linearly between each syntactic head and its dependent (excluding punctuation dependencies). \\
LFTK \shortcite{lee-lee-2023-lftk} &   \texttt{n\_adp}  & 0.4144 & Number of adpositions. \\
LingFeat \shortcite{lee-etal-2021-pushing} &   \texttt{SquaAjV\_S}  & 0.4088 & Squared Adjective Variation-1, which is calculated as the $( \frac{(\text{number-of-unique-adjectives})^2} { \text{number-of-total-adjectives}} )$.\\
LFTK \shortcite{lee-lee-2023-lftk} &   \texttt{n\_upunct}  & 0.4053 & Number of unique punctuations. \\
LFTK \shortcite{lee-lee-2023-lftk} &   \texttt{corr\_adp\_var}  & 0.4031 & Corrected adposition variation, which is computed as $(\frac{\text{number-of-unique-adpositions}} { \sqrt{2 \times \text{number-of-all-adpositions}}})$  \\
LFTK \shortcite{lee-lee-2023-lftk} &   \texttt{n\_uadp}  & 0.4022 & Number of unique adpositions. \\
LFTK \shortcite{lee-lee-2023-lftk} &   \texttt{corr\_propn\_var}  & 0.3895 & Corrected proper noun variation, which is computed as  $(\frac{\text{number-of-unique-proper-nouns} }{ \sqrt{2 \times \text{number-of-all-proper-nouns}}})$\\
LingFeat \shortcite{lee-etal-2021-pushing} &   \texttt{WClar20\_S}  & 0.3879 & Semantic Clarity measured by averaging the differences between the primary topic's probability and that of each subsequent topic, reflecting how prominently a text focuses on its main topic, based on 200 topics extracted from Wikipedia Corpus. \\
LingFeat \shortcite{lee-etal-2021-pushing} &   \texttt{SquaNoV\_S}  & 0.3864 & Squared Noun Variation-1, which is calculated as the $((\text{number-of-unique-nouns})^2 / \text{number-of-total-nouns} )$.\\
\bottomrule
\end{tabular}
}

\label{table:appendix-all-features}
\captionof{table}{Top 50 most influential linguistic features on readability assessment (continue). }
\end{minipage}

\clearpage

\section{Introduction of  Medical Text Simplification Resources}
\label{appendix:intro-resources}

Our dataset is constructed on top of open-accessed resources. Each of the resources is detailed below. Table \ref{table:venue} presents the basic statistics of 180 sampled article (segment) pairs.

\paragraph{Biomedical Journals.} The latest advancements in the medical field are documented in the research papers. To improve accessibility, the authors or domain experts sometimes write  a summary in lay language, providing a valuable resource for studying medical text simplification.  We include five sub-journals from NIHR, five sub-journals from PLOS, and the Proceedings  of the National Academy of Sciences (PNAS) compiled by \cite{guo2022cells}.  In addition, we also include the eLife corpus compiled by \cite{goldsack-etal-2022-making}, which consists of the paper abstracts and  summaries in life sciences written by expert editors.

\paragraph{Cochrane Reviews.} 
As ``the highest standard in evidence-based healthcare'', Cochrane Review\footnote{\url{https://www.cochranelibrary.com/}} provides  systematic reviews for the effectiveness of interventions and the quality of  diagnostic tests in healthcare and health policy areas, by identifying, appraising, and synthesizing all the empirical evidence that meets pre-specified eligibility criteria. We use the parallel corpus compiled by \cite{devaraj-etal-2021-paragraph}.

\paragraph{Medical Wikipedia.} As their original and simplified versions are created independently in a collaboration process, the two versions  are on the same topic but may  not be entirely aligned \cite{xu-etal-2015-problems}. We apply the state-of-the-art methods \cite{jiang-etal-2020-neural} to extract aligned paragraph pairs from Wikipedia, of which we improve the quality and quantity over existing work \cite{pattisapu2020leveraging}. 
Specifically, we first collect 60,838 medical terms using Wikidata's SPARQL service\footnote{\url{https://query.wikidata.org/}} by querying unique terms that have 30 specific properties, including UMLS code, medical encyclopedia, and the ontologies for disease, symptoms, examination, drug, and therapy.
Then, we extract corresponding articles for each term from Wikipedia and simple Wikipedia dumps,\footnote{The March 22, 2023 version.} based on title matching using WikiExtractor library,\footnote{\url{https://attardi.github.io/wikiextractor/}} resulting in 2,823 aligned article pairs after filtering the empty pages.   Finally, we use the state-of-the-art neural CRF sentence alignment model \cite{jiang-etal-2020-neural} with 89.4 F1 on Wikipedia to perform paragraph and sentence alignment for each complex-simple article pair. 

\paragraph{Merck Manuals.} We use the  segment pairs from prior work \cite{cao-etal-2020-expertise}, which are manually aligned by medical experts.

\renewcommand{\arraystretch}{1.15}
\begin{table}[t!]
\small
\centering
\resizebox{\linewidth}{!}{
\begin{tabular}{@{\hspace{0.02cm}}L{5.29cm}@{\hspace{-0.15cm}}C{1.9cm}@{\hspace{-0.1cm}}C{2.1cm}@{\hspace{0.02cm}}}
\toprule
\multirow{2}{*}{\textbf{Source of the Publication}} & \textbf{Avg. \#Sent.} &  \textbf{Avg. Sent. Len.}    \\
& Comp./Simp. & Comp./Simp. \\
\midrule

\multicolumn{3}{@{\hspace{0.02cm}}l}{\textbf{\textit{Public Library of Science (PLOS)}}} \\ 
\midrule
Biology & 8.3 / 8.2 & 28.2 / 26.8 \\
Genetics & 10.2 / 6.2 & 28.9 / 30.3 \\
Pathogens & 8.9 / 7.2 & 30.7 / 29.5 \\
Computational Biology & 9.1 / 7.2 & 29.3 / 27.4 \\
Neglected Tropical Diseases  & 10.2 / 8.0 & 29.3 / 26.4 \\
\midrule
\multicolumn{3}{@{\hspace{0.02cm}}l}{\textbf{\textit{National Institute for Health and Care Research (NIHR)}}} \\ 
\midrule
Public Health Research & 23.4 / 14.3 & 26.2 / 20.5 \\
Health Technology Assessment & 25.1 / 12.9 & 27.3 / 25.7 \\
Efficacy and Mechanism Evaluation & 22.6 / 14.9 & 28.2 / 21.4 \\
Programme Grants for Applied Research & 27.6 / 14.2 & 27.6 / 22.6 \\
Health Services and Delivery Research & 23.2 / 14.1 & 27.9 / 23.2 \\
\midrule 
 Medical Wikipedia & 5.4 / 5.8 & 23.3 / 19.4 \\
 Merck Manuals (medical references)  &  5.0 / 5.6 & 23.8 / 16.3 \\
eLife (biomedicine and life sciences) & 6.5 / 15.6 & 27.0 / 26.3 \\
Cochrane Database of Systematic Reviews &  25.4 / 16.1 & 27.3 / 22.2 \\
Proc. of  National Academy of Sciences & 9.1 / 5.5 & 27.2 / 24.1 \\

\bottomrule
\end{tabular}
}
\vspace{-5pt}
\caption{Average \# of sentences and their length for 180 sampled parallel articles (segments) from 15 resources.}
\label{table:venue}
\vspace{-15pt}
\end{table}

\section{Implementation Details for Complex Span Identification Models}
\label{appendix:cwi_iumplementation_details}
We use the Huggingface\footnote{\url{https://github.com/huggingface/transformers}} implementations of the BERT and RoBERTa models.  We tune the learning rate in \{1e-6, 2e-6, 5e-6, 1e-5, 2e-5\} based on F1 on the devset, and find 2e-6 works best for our best performing  RoBERTa-large model. All models are trained  within 1.5 hours on one NVIDIA A40 GPU. 

\section{More Related work on Complex Span Identification in Medical Domain}
\label{sec:more-related-work}
Other work mainly focuses on the general domains such as news and Wikipedia, including CW corpus in SemEval 2016 shared task \cite{shardlow-2013-cw, paetzold-specia-2016-semeval} and CWIG3G2 corpus in SemEval 2018 \cite{yimam-etal-2017-cwig3g2, yimam-etal-2018-report}. In addition, \citet{guo2023personalized} collects a jargon dataset from computer science research papers, \citet{lucy-etal-2023-words} studies the social implications of jargon usage, and \citet{august-etal-2022-generating,huang-etal-2022-understanding} focus on the explanation of jargon.

\clearpage
\section{More Results for Complex Span Identification}
\label{appendix:cwi_imore_performance} 

Table \ref{table:complex-span-identification-em-entity}  presents the results of the exact match at entity level for the complex span identification task on the {\medcsi} test set. As medical jargon and complex spans have  diverse formats in the medical articles, it is challenging for the models to predict the exact matched entities.

\begin{table}[pht!]
\renewcommand{\arraystretch}{1.2}
\setlength\tabcolsep{2pt} %

\small
\centering

\begin{tabular}{lccc}
\toprule

Models  &  Binary   & 3-Class & 7-Category   \\

    \midrule

    \multicolumn{3}{l}{\textbf{\textit{Large-size Models}}}\\ [-0.1ex]
    \midrule

BERT \shortcite{devlin-etal-2019-bert} & 72.0 & 68.2 & 48.5 \\
RoBERTa \shortcite{liu2019roberta} & \textbf{74.9} & \textbf{71.2} & \textbf{64.1} \\
BioBERT \shortcite{lee2020biobert} & 72.4 & 67.6 & 60.5 \\
PubMedBERT \shortcite{pubmedbert} & 73.4 & 69.9 & 62.2 \\

    \midrule

    \multicolumn{3}{l}{\textbf{\textit{Base-size Models}}}\\ [-0.1ex]
    \midrule
    
BERT \shortcite{devlin-etal-2019-bert} & 70.7 & 67.0 & 59.3 \\
RoBERTa \shortcite{liu2019roberta} & \underline{73.5} & \underline{70.0} & \underline{62.4} \\
BioBERT \shortcite{lee2020biobert} & 70.5 & 67.1 & 59.8 \\
PubMedBERT \shortcite{pubmedbert} & 72.2 & 69.0 & 61.2 \\

    \bottomrule
\end{tabular}
\caption{\textbf{Micro F1} of exact match at entity-level for complex span identification task on the {\medcsi} test set. The \textbf{best} and \underline{second best} scores within each model size are highlighted. Models are trained with fine-grained labels in seven categories and evaluated at different granularity.}
\label{table:complex-span-identification-em-entity}
\vspace{-15pt}
\end{table}

\section{More Results on Medical Readability Prediction}

We conducted an additional experiment to study how different complex span identification models used in Section \ref{sec:cwi_experiment} affect the performance of medical readability prediction. We find that using predictions from different complex span prediction models leads to similar improvements in readability prediction, with a $\pm$ 0.015 difference in average Pearson correlation across different resources.

\clearpage
\onecolumn

\section{Prompts for Sentence Readability}
\label{appendix:readability-prompt-template}

\noindent\begin{minipage}{\textwidth}
\centering
\begin{tabular}{@{}p{\linewidth}@{}}
\toprule
Rate the following sentence on its readability level. The readability is defined as the cognitive load required to understand the meaning of the sentence. Rate the readability on a scale from very easy to very hard. Base your scores on the CEFR scale for L2 learners. You should use the following key:\\ 
1 = Can understand very short, simple texts a single phrase at a time, picking up familiar names, words and basic phrases and rereading as required.\\
2 = Can understand short, simple texts on familiar matters of a concrete type\\
3 = Can read straightforward factual texts on subjects related to his/her field and interest with a satisfactory level of comprehension.\\
4 = Can read with a large degree of independence, adapting style and speed of reading to different texts and purpose\\
5 = Can understand in detail lengthy, complex texts, whether or not they relate to his/her own area of speciality, provided he/she can reread difficult sections.\\
6 = Can understand and interpret critically virtually all forms of the written language including abstract, structurally complex, or highly colloquial literary and non-literary writings.\\
EXAMPLES:\\
Sentence: ``[EXAMPLE 1]''\\
Given the above key, the readability of the sentence is (scale=1-6): [RATING 1]\\
\addlinespace
Sentence: ``[EXAMPLE 2]''\\
Given the above key, the readability of the sentence is (scale=1-6): [RATING 2]\\
\addlinespace
Sentence: ``[EXAMPLE 3]''\\
Given the above key, the readability of the sentence is (scale=1-6): [RATING 3]\\
\addlinespace
Sentence: ``[EXAMPLE 4]''\\
Given the above key, the readability of the sentence is (scale=1-6): [RATING 4]\\
\addlinespace
Sentence: ``[EXAMPLE 5]''\\
Given the above key, the readability of the sentence is (scale=1-6): [RATING 5]\\
\addlinespace
Sentence: ``[TARGET SENTENCE]''\\
Given the above key, the readability of the sentence is (scale=1-6): [RATING]\\
\bottomrule
\end{tabular}
\vspace{-5pt}
\captionof{table}{Following \cite{naous2023towards} in prompt construction, we utilize the same description of the six CEFR levels that were provided to human annotators, along with five examples and their ratings, randomly sampled from the dev set. Then, the model is instructed to evaluate the readability of a given sentence. The full template is presented above.}
\end{minipage}

\clearpage
\onecolumn
\section{Annotated Screenshot of Search Engine Results}
\label{appendix:annotated-google}

\noindent\begin{minipage}{\textwidth}
   \small
    \centering
    \includegraphics[width=\textwidth]{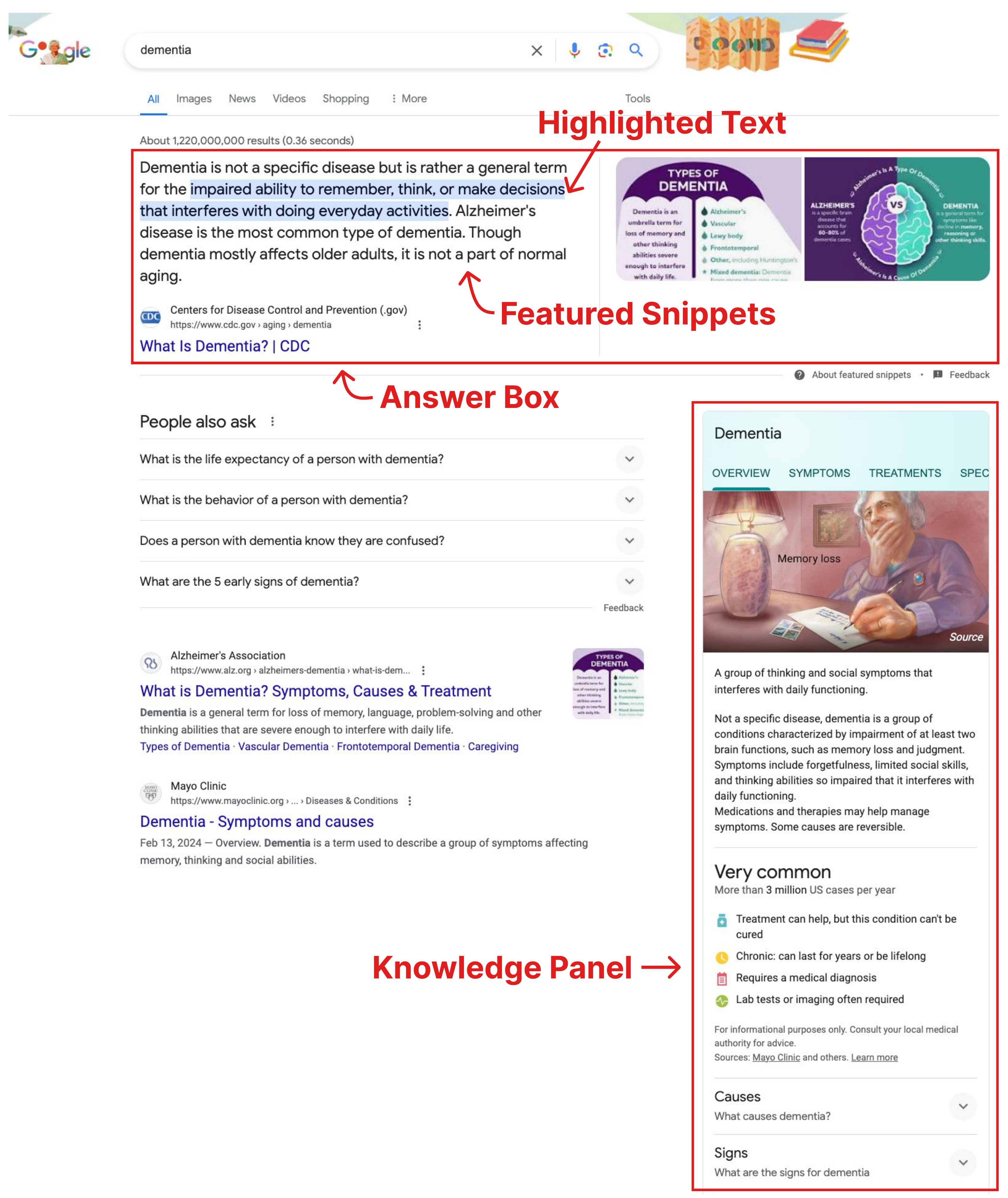}
    \captionof{figure}{An annotated screenshot of search results from Google. Search engines may provide the explanation of a medical term in two places: (1) the feature snippets in the answer box and (2) the knowledge panel on the right-hand side, which is powered by a knowledge graph.}
    \label{fig:annotated-google-screenshot}
\end{minipage}

\clearpage
\onecolumn
\section{Annotation Interface for Sentence Readability}
\label{appendix:annotation-interface}

\noindent\begin{minipage}{\textwidth}
   \small
    \centering
    \includegraphics[width=\textwidth]{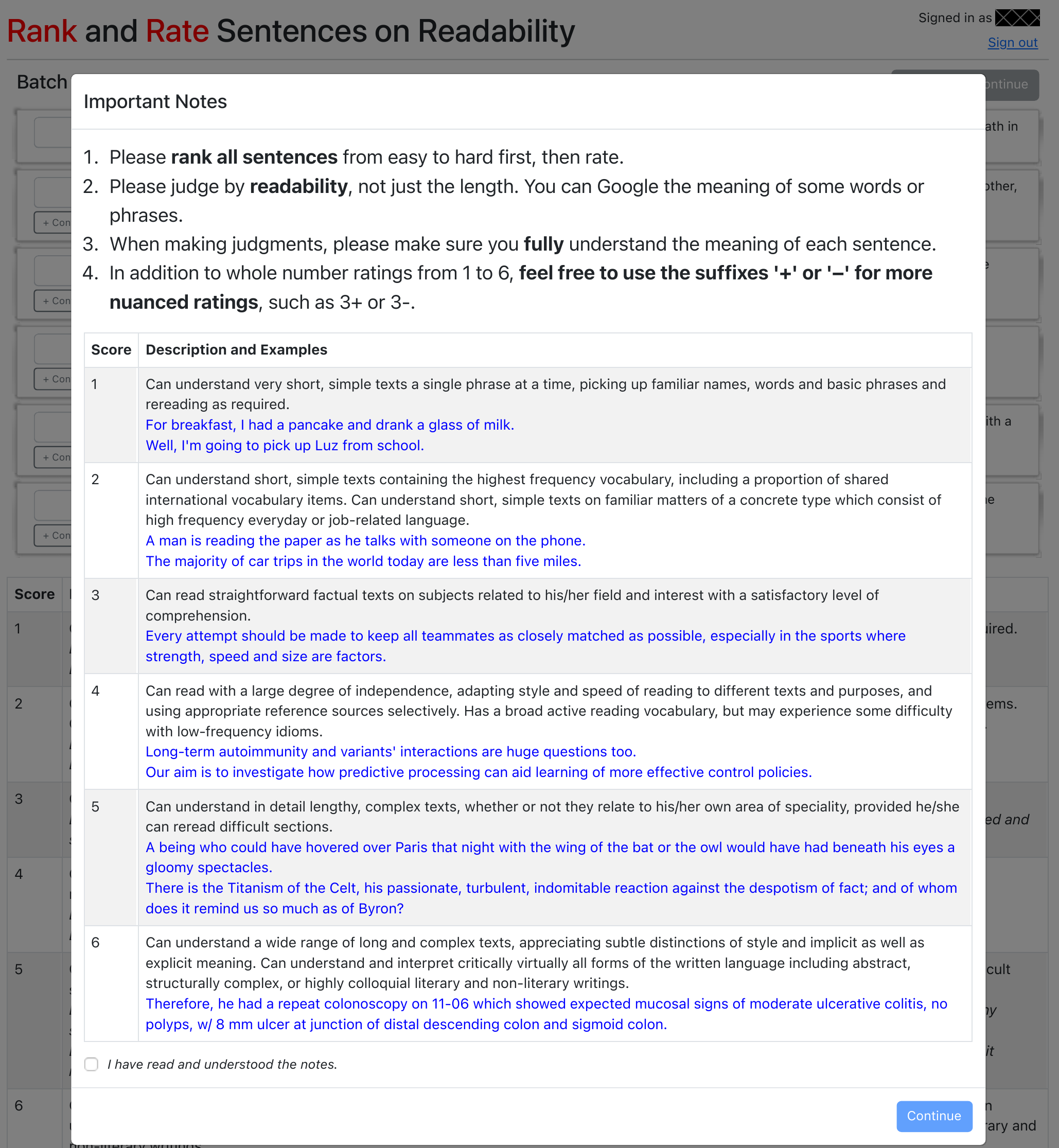}
    \captionof{figure}{Instructions for annotating the sentence readability.}
\end{minipage}

\clearpage
\noindent\begin{minipage}{\textwidth}
   \small
    \centering
    \includegraphics[width=\textwidth]{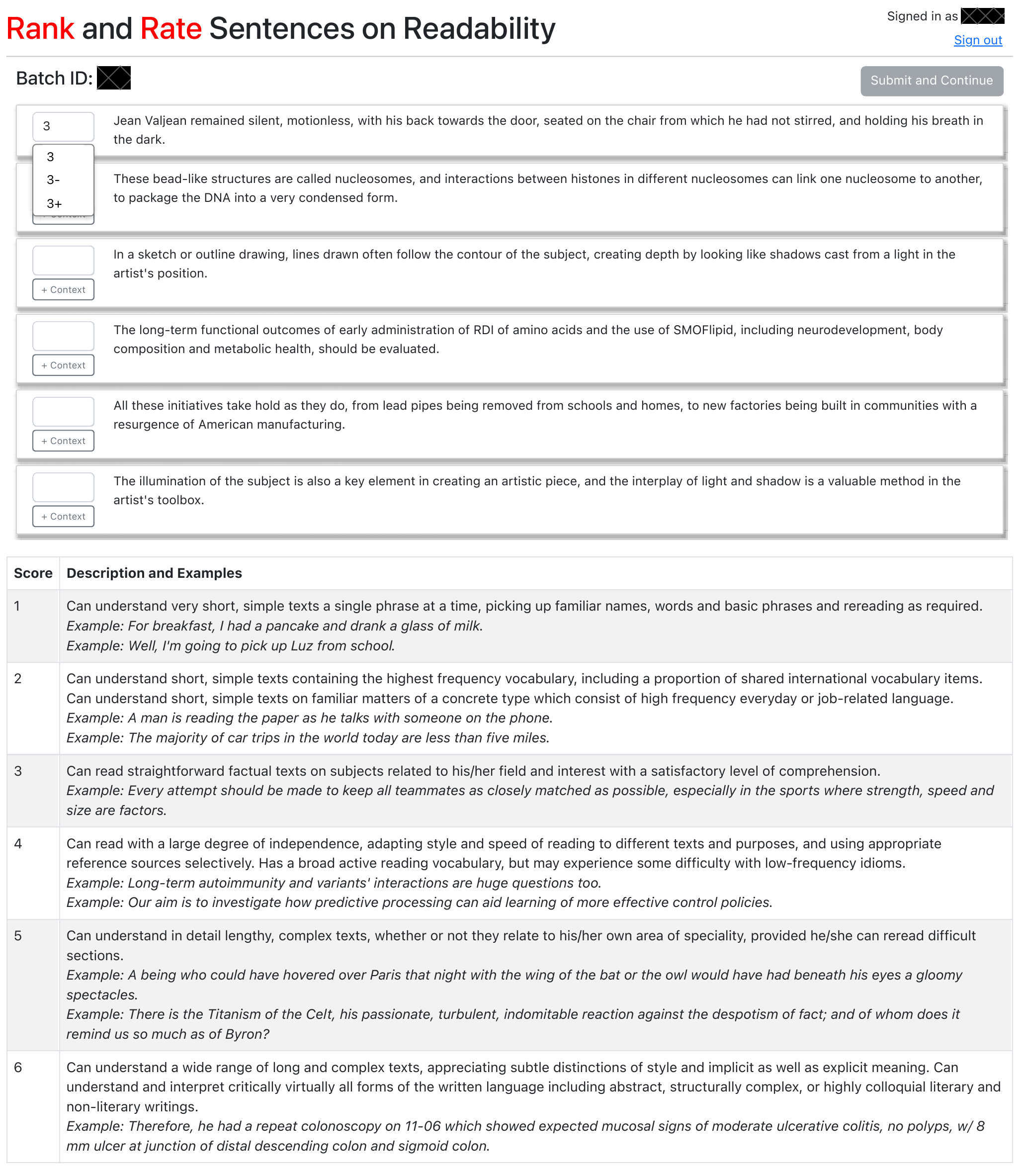}
    \captionof{figure}{The interface for annotating sentence readability. Annotators can click the ``+ Context'' button to see the surrounding sentences.}
\end{minipage}

\clearpage
\section{Annotation Interface for Complex Span Identification}
\label{section:span-annotation}
\vspace{+10pt}
\noindent\begin{minipage}{\textwidth}
   \small
    \centering
    \includegraphics[width=\textwidth]{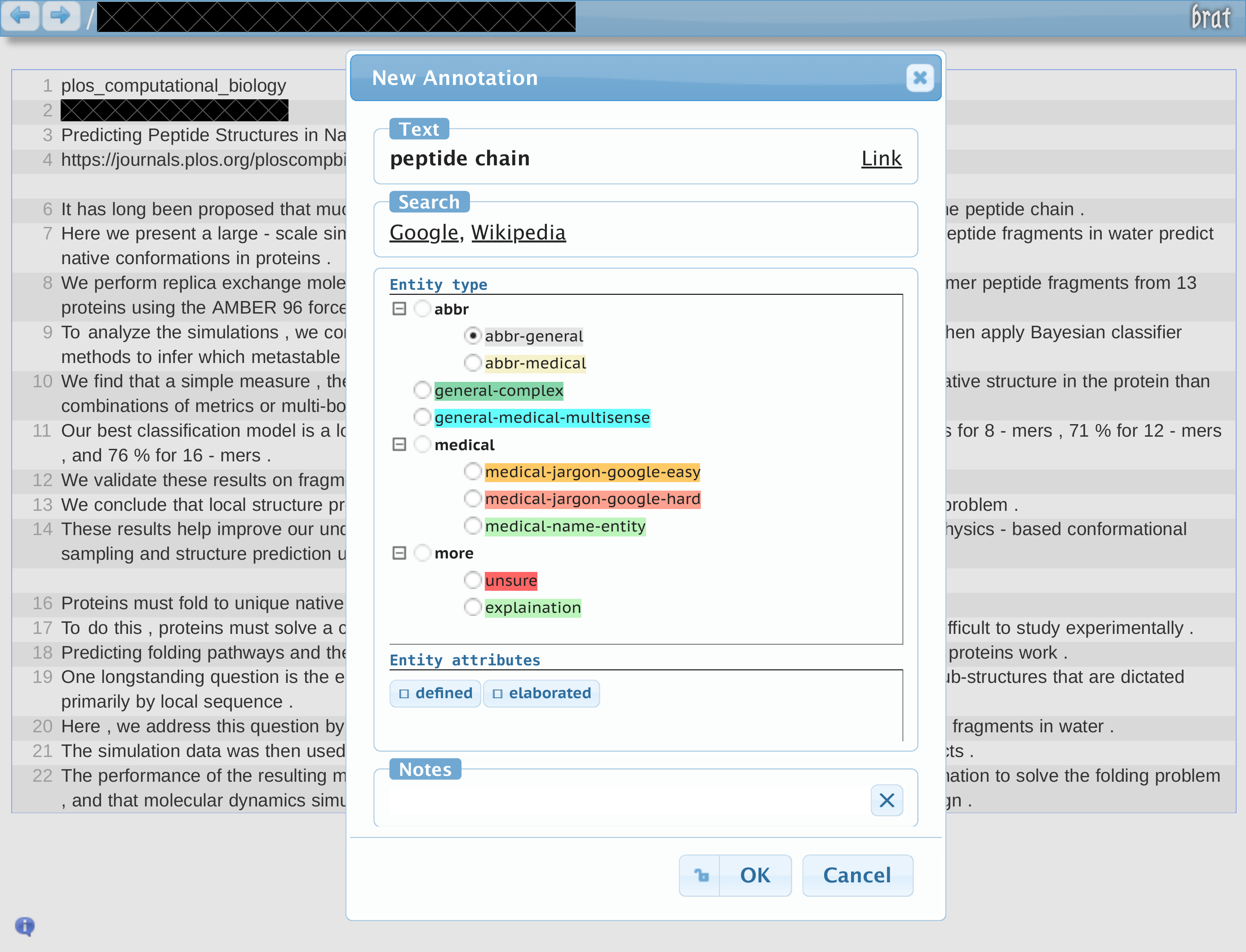}
    \captionof{figure}{The annotation interface for complex span identification.}
\end{minipage}

\clearpage
\section{Annotation Guideline for Complex Span Identification}
\label{section:span-annotation-guideline}
\vspace{+10pt}
\noindent\begin{minipage}{\textwidth}
   \small
    \centering
    \fbox{\includegraphics[width=\textwidth, clip, trim=2cm 2cm 2cm 2cm]{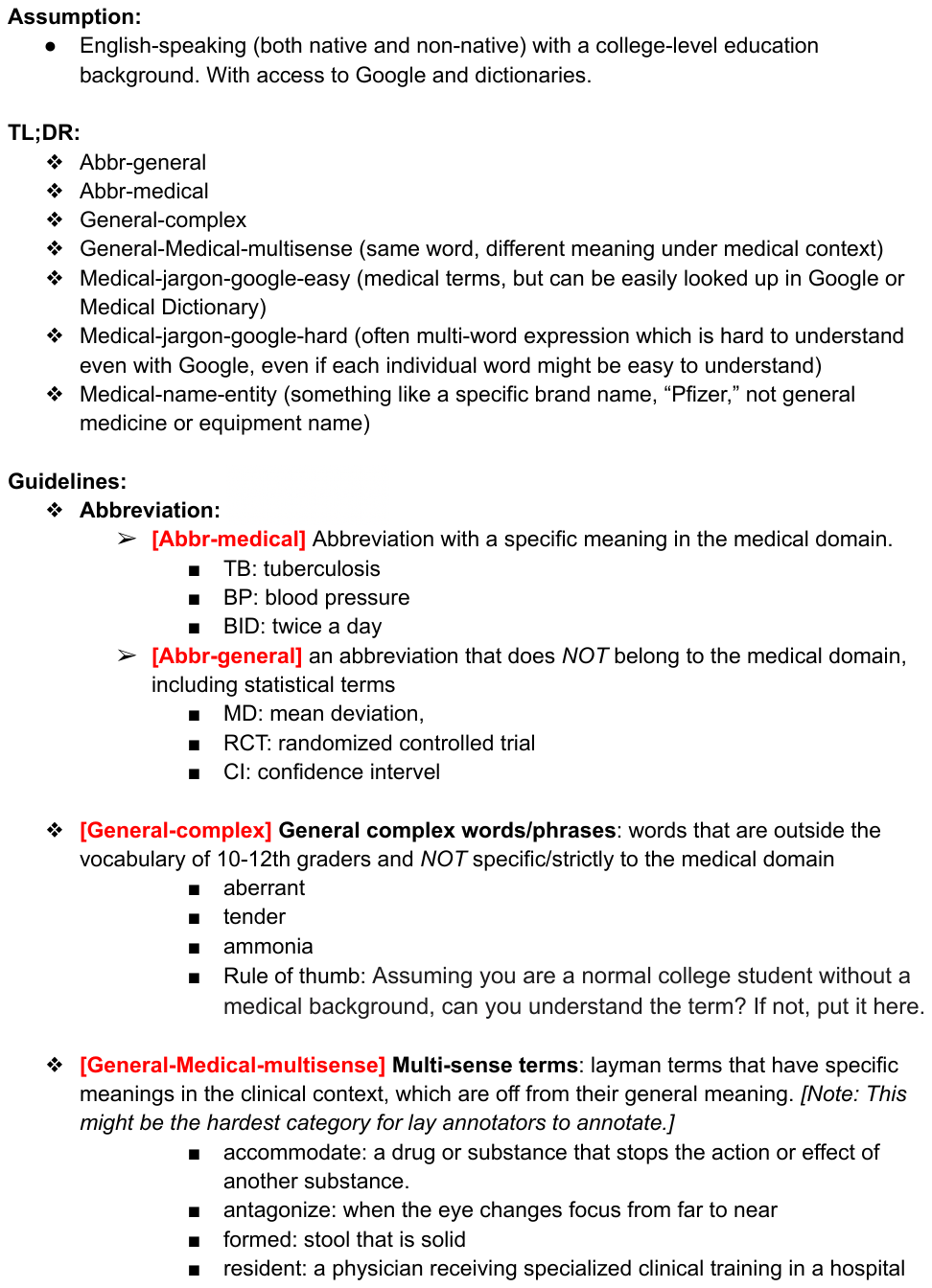}}
    \captionof{figure}{The annotation guideline for complex span identification.}
\end{minipage}

\clearpage

\noindent\begin{minipage}{\textwidth}
   \small
    \centering
    \fbox{\includegraphics[width=\textwidth, clip, trim=2cm 2cm 2cm 2cm]{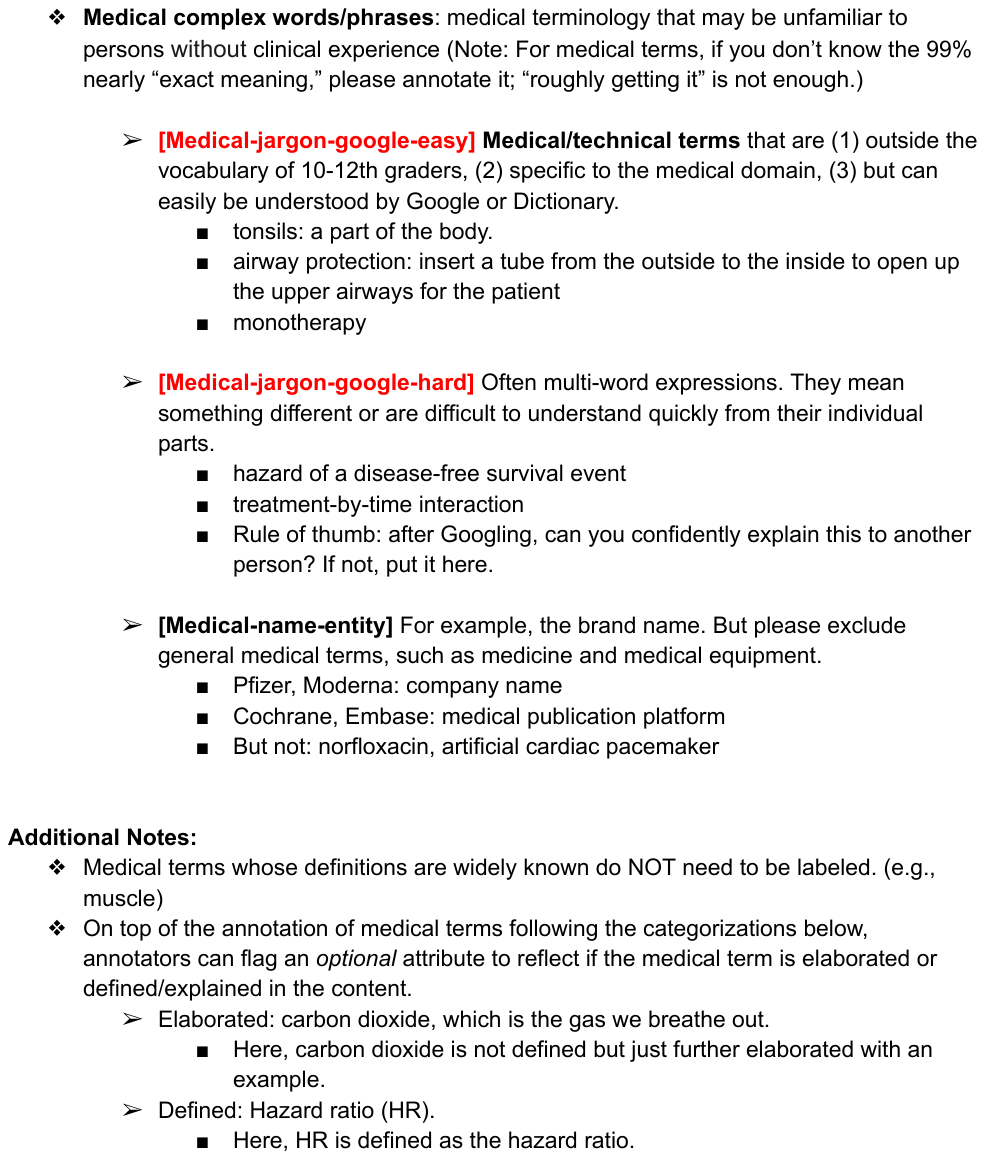}}
    \captionof{figure}{The annotation guideline for complex span identification (continue).}
\end{minipage}

\end{document}